\documentclass[lettersize,journal]{IEEEtran}
\usepackage{amsmath,amsfonts}
\usepackage{algorithmic}
\usepackage{algorithm}
\usepackage{array}
\usepackage{textcomp}
\usepackage{stfloats}
\usepackage{url}
\usepackage{verbatim}
\usepackage{cite}
\usepackage{tabularx} 
\usepackage{booktabs} 
\usepackage{graphicx} 
\usepackage[caption=false,font=normalsize,labelfont=sf,textfont=sf]{subfig}

\begin{document}

\title{Controllable Risk Scenario Generation from Human Crash Data for Autonomous Vehicle Testing}

\author{Qiujing Lu, Xuanhan Wang, Runze Yuan, Wei Lu, Xinyi Gong, Shuo Feng$^*$
\thanks{This work is supported by Beijing Natural Science Foundation (L243025), National Natural Science Foundation of China (No.62473224) and CCF-DiDi GAIA Collaborative Research Funds. (Corresponding author: Shuo Feng.)}
\thanks{Q. Lu, X. Wang, R. Yuan and S. Feng are with Department of Automation, Tsinghua University, Beijing, 100084, China  (E-mail address: fshuo@tsinghua.edu.cn). 
W. Lu is with Beijing DiDi Infinity Technology and Development Co., Ltd., Beijing, 100095, China. 
X. Gong is with Space Information Research Institute and Zhejiang Key Laboratory of Space Information Sensing and Transmission, Hangzhou Dianzi University, Hangzhou, 310018, China.}
}

\markboth{IEEE TRANSACTIONS ON INTELLIGENT TRANSPORTATION SYSTEMS,~Vol.~, No.~, Month~2025}%
{Shell \MakeLowercase{\textit{et al.}}: A Sample Article Using IEEEtran.cls for IEEE Journals}

\IEEEpubid{0000--0000/00\$00.00~\copyright~2021 IEEE}

\maketitle

\begin{abstract}
Ensuring the safety of autonomous vehicles (AV) requires rigorous testing under both everyday driving and rare, safety-critical conditions. A key challenge lies in simulating environment agents, including background vehicles (BVs) and vulnerable road users (VRUs), that behave realistically in nominal traffic while also exhibiting risk-prone behaviors consistent with real-world accidents. We introduce Controllable Risk Agent Generation (CRAG), a framework designed to unify the modeling of dominant nominal behaviors and rare safety-critical behaviors. CRAG constructs a structured latent space that disentangles normal and risk-related behaviors, enabling efficient use of limited crash data. By combining risk-aware latent representations with optimization-based mode-transition mechanisms, the framework allows agents to shift smoothly and plausibly from safe to risk states over extended horizons, while maintaining high fidelity in both regimes. Extensive experiments show that CRAG improves diversity compared to existing baselines, while also enabling controllable generation of risk scenarios for targeted and efficient evaluation of AV robustness.
\end{abstract}


\section{Introduction}
\IEEEPARstart{T}{he} advent of autonomous vehicles (AVs) represents a transformative shift in transportation, with the potential to enhance mobility, improve traffic efficiency, and reduce accidents. Despite the promise of advanced AV technology, numerous corner cases observed in real-world deployments have highlighted the safety limitations of current systems. Identifying and addressing these corner cases is a critical step toward overcoming the bottlenecks hindering large-scale deployment. 

Given the high costs, inefficiencies, and lack of repeatability and controllability associated with road testing, realistic simulation-based testing emerges as one of the most promising approaches for advancing AV safety and reliability. Nevertheless, leveraging simulation as a viable tool for AV development and testing remains challenging. It requires a simulated environment populated with agents that are both realistic and responsive. Human driving behavior, however, is highly unpredictable, shaped by diverse intentions and varying driving styles. This complexity poses a significant scientific challenge for constructing generative models that can accurately reproduce such behaviors. Although large-scale driving datasets collected by autonomous driving companies \cite{mei2022waymo,caesar2020nuscenes} have facilitated the modeling of nominal driving behavior, safety-critical scenarios remain extremely rare. Furthermore, as highlighted by \cite{liu2024curse}, improving safety performance paradoxically makes it harder to learn safety-critical behavior. The safer an autonomous system becomes, the fewer dangerous situations it encounters, and thus the data required to learn how to handle such events becomes even more scarce. This leads to a severe class imbalance between abundant normal driving samples and scarce high-risk samples, making purely data-driven models insufficient to capture both regimes faithfully.

Most prior work tackles these regimes in isolation: imitating nominal human driving or synthesizing risky behaviors. It overlooks the need for a unified model that explicitly captures the shift from nominal to risky behavior. Yet modeling the temporal evolution of safety-critical scenarios is essential for AV testing, as it reveals how driving policies may precipitate or mitigate incidents. Progress toward this goal is further hindered by the scarcity and diversity of safety-critical events, which limits the effectiveness of purely data-driven methods.

In view of these challenges, it is noteworthy that although AV driving logs contain a small number of real accident cases, large-scale human-driven accident data continue to accumulate. How to efficiently leverage human crash data, to generate risk scenarios for AV testing remains an open and pressing challenge.

\begin{figure}[h]
  \centering
  \includegraphics[width=\columnwidth]{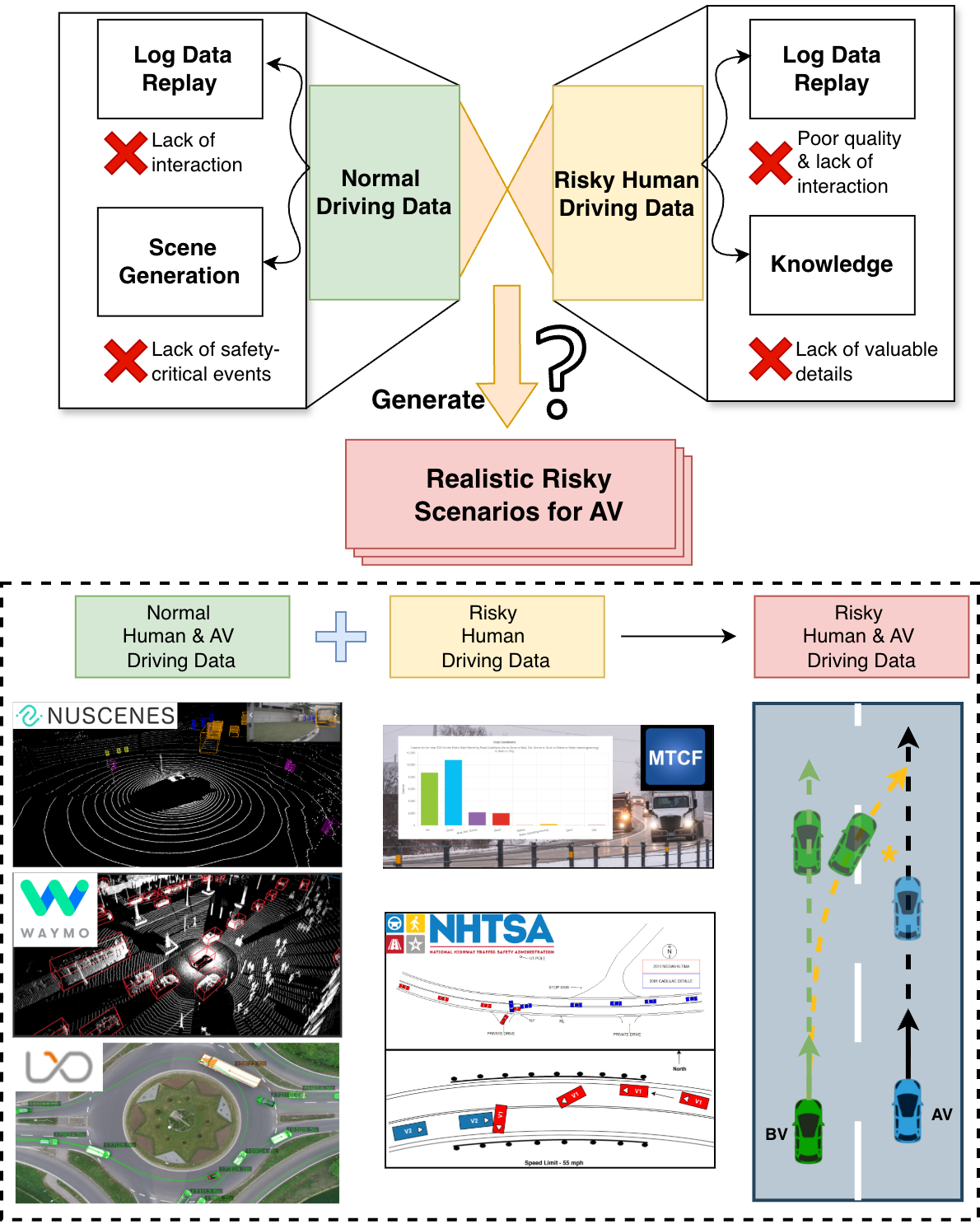}
   \caption{Traditional AV testing isolates normal and risky driving data, hindering the generation of safety-critical scenarios in simulation. CRAG leverages both data types to enhance the realism and quality of generated risky scenarios.}
  \label{fig:problem_intro}
\end{figure}

\IEEEpubidadjcol

To address these challenges, this study introduces a novel framework that bridges normal driving data with limited human accident data to enable AV testing under realistic safety-critical conditions. It efficiently learns from scarce accident data by constructing a compact latent risk space using unsupervised learning. Realistic risky motions are then generated by integrating latent risk factors and motion knowledge from nominal behaviors through optimization-guided state search. The proposed framework facilitates the generation of full-spectrum scenarios, providing a controllable environment for evaluating AV performance under diverse risk conditions.

Our contribution includes:
\begin{itemize}
\item Presenting a framework that leverages human driving accident data to transfer risk knowledge for AV testing with fine-grained control over risk scenario types.
\item Learning compact and transferable risk representations within a structured latent space derived from limited accident data.  
\item Efficiently searching for and synthesizing targeted risk states from limited accident evidence using a normal motion predictor guided by latent representations.  
\end{itemize}

The remainder of this paper is structured as follows: Section \ref{sec:Related work} presents a review of relevant literature. Section \ref{sec:framework} details the methodology for implementing the model. In Section \ref{sec:setup}, the complete pipeline and evaluation methods are described. Section \ref{sec:exp} discusses the simulation results, and finally, Section \ref{sec:conclusion} provides the conclusion.

\section{Related work}
\label{sec:Related work}
\subsection{Simulation-Based Testing for AV}
Simulation-based testing plays a crucial role in AV testing, supporting large-scale, repeatable evaluations of driving behavior \cite{chen2024data}. Nevertheless, even with abundant driving data, constructing high-value simulated scenarios for AV safety assessment remains challenging.

Replaying logged trajectories for all agents preserves high motion fidelity \cite{arief2018accelerated,montali2023waymo,li2024scenarionet} yet such replay is inherently non-reactive. The environment cannot adapt to novel AV behaviors, so any deviation from the recorded trajectories breaks interaction fidelity, reducing reactivity, diversity, and counterfactual coverage—and thus weakening evaluation validity \cite{liu2024towards}. In parallel, work such as \cite{yan2024street} and \cite{guo2023streetsurf} advances scene generation through photorealistic novel-view synthesis for dynamic urban environments. These approaches markedly improve visual fidelity but treat agent decision-making as largely exogenous, thereby leaving the gap in interaction realism unresolved.

Closed-loop evaluation, by contrast, requires BV behavior models that respond realistically to both the AV and surrounding traffic \cite{yan2023learning,ren2025intelligent}. Such models are essential for assessing whether the ego vehicle interacts safely with its environment, thereby supporting effective and reliable AV testing. However, constructing BV models that remain stable over long horizons, generalize to previously unseen states, and adapt plausibly to the ego vehicle’s evolving policy remains technically challenging.

\subsection{Scenario Generation with Simulated Agents}
To model realistic and reactive simulated agents that can interact plausibly with surrounding traffic, three mainstream approaches are commonly employed: rule-based, statistical-based, and data-driven learning-based. Many previous studies have achieved agents' behavior generation using rule-based or statistical-based methods. Traditional rule-based models, such as the Intelligent Driver Model (IDM) \cite{Treiber_Hennecke_Helbing_2000} and MOBIL \cite{Kesting_Treiber_Helbing_2007} are overly accommodating and lack the diversity of human driving behaviors on a micro level. In addition, these methods may suffer from limited accuracy, poor generalization capability, a lack of adaptive updating, and the need for significant expert knowledge to establish rules \cite{chen2024data}. Some statistical methods \cite{laugier2011probabilistic,chen2020driving,aoude2012driver} employ probabilistic models to characterize the patterns and distributions of agent behaviors. These models can then be used to predict future motion sequences for generating testing scenarios. However, they typically assume independence among agents, overlooking the complex and subtle interactions that occur in traffic. As a result, the simulated agents may behave unrealistically, with limited or no natural interactions with AVs.

Compared with the above methods, data-driven deep learning models are more promising considering their adaptability and capacity to learn complex patterns. TrafficSim \cite{Suo_Regalado_Casas_Urtasun_2021} employs a GRU \cite{cho2014learning} combined with CNN encoders to imitate human driving behaviors from real-world data. TrafficGen \cite{Feng_Li_Peng_Tan_Zhou_2023} adopts an autoregressive neural generative model, while  \cite{Wang_Zhao_Yi_2023} leverage a Transformer architecture with a receding-horizon prediction mechanism to enhance realism. Symphony \cite{Igl_Kim_Kuefler_Mougin_Shah_Shiarlis_Anguelov_Palatucci_White_Whiteson_2022} integrates learned policies with a parallel beam-search procedure, further improving the plausibility and diversity of generated behaviors. Overall, these approaches exhibit strong generalization in modeling human-like interactions across diverse traffic environments.
Recent efforts have also explored building reactive simulator and  benchmarks for closed-loop evaluation, such as nuPlan \cite{caesar2021nuplan}, Nocturne \cite{vinitsky2022nocturne}, MetaDrive \cite{li2022metadrive}, and Waymax \cite{gulino2024waymax}. Yet, these framework devote limited attention to ensuring diversity and realism in rare safety-critical interactions. Most existing methods emphasize imitation of normal human driving behaviors and are trained or evaluated on naturalistic datasets that seldom contain accident events. This restriction inherently limits their ability to generate realistic and diverse high-risk testing scenarios for safety evaluation. Consequently, creating realistic safety-critical scenarios in closed-loop simulation remains an underexplored and technically challenging problem.

\subsection{Safety-Critical Scenario Generation}
Simulating the rare but consequential events is essential for reliable testing of AVs under extreme conditions \cite{fremont2023scenic,Ding_Xu_Arief_Lin_Li_Zhao_2023,luracl}. Some approaches exploit prior knowledge. For example, \cite{Rana_Malhi_2021} use parametric equations to construct random risk scenarios, while \cite{Althoff_Lutz_2018} optimize the motion of surrounding vehicles within constrained drivable areas. More recently, multimodal large language models have been applied to reason about risky situations and generate scenarios with semantic guidance \cite{aasi2024generating,lu2024realistic,lu2024multimodal}. Other methods adopt a data-driven perspective. For instance, \cite{Scanlon_Kusano_Daniel_Alderson_Ogle_Victor_2021} reconstructs fatal accidents by perturbing key parameters from real-world logs to synthesize counterfactual variations.

Adversarial techniques have also been applied to elicit risky situations by deliberately challenging AV systems. These methods push vehicles into corner cases that stress decision-making and control policies. \cite{Wang_Pun_Tu_Manivasagam_Sadat_Casas_Ren_Urtasun_2021} generates adversarial agent behaviors and modifies LiDAR sensor inputs accordingly to expose vulnerabilities in perception and planning. 
\cite{Rempe_Philion_Guibas_Fidler_Litany_2022} and \cite{zhang2023cat} employ traffic prior, to optimizing agents' trajectories to induce collisions. \cite{Hanselmann_Renz_Chitta_Bhattacharyya_Geiger_2022} leverages kinematic gradients to discover adversarial behaviors that directly exploit weaknesses of real AV systems. More recently, diffusion-based approaches incorporate collision-oriented cost functions during inference to promote adversarial agent generation \cite{Zhong_Rempe_Xu_Chen_Veer_Che_Ray_Pavone_2023,chang2023controllable}.  While effective for stress-testing, these methods often depend on altered initial conditions or hand-crafted objectives, which can reduce the naturalness and validity of the simulated scenarios.

These limitations highlight the need for more realistic modeling of adversarial and interactive agent behaviors. The primary challenge arises from the scarcity of AV accident data, which hinders direct learning of risky behaviors. A key missing component is an effective mechanism to transfer risk-related factors from existing human accident data into simulated environments. Furthermore, maintaining model robustness over long simulation horizons while preserving realism, reactivity, and diversity in scenario generation remains an open challenge \cite{yan2023learning,feng2023dense}.

\section{Methodology}
\label{sec:framework}
In this section, we first formalize the problem of closed-loop risk scenario generation. We then introduce a risky latent space constructed using a variational autoencoder (VAE) trained on crash data. Finally, we propose an optimization-based method to search for risky states within this latent space, enabling the generation of controllable safety-critical scenarios.

\subsection{Problem Formulation}
In closed-loop simulation, the generated behaviors of agents at time $t$ are fed back as updated inputs for generating their decisions at the next time step. To formalize this process, we first define the agent states. The state of agent $i$ at time $t$ is denoted as $\boldsymbol{s}_i^t$, includes its two-dimensional center position ($x_i^t$, $y_i^t$), heading $h_i^t$ and velocity $v_i^t$. The joint state of all agents at time $t$ is $\boldsymbol{S}^t = [\boldsymbol{s}_1^t, \boldsymbol{s}_2^t, \ldots, \boldsymbol{s}_N^t]$. 

The key component enabling this process is the motion predictor, which maps historical information to future agent behaviors. Without loss of generality, let $D(\theta)$ denote a motion predictor parameterized by $\theta$, which predicts multi-modal future trajectories for each agent in the scenario. Formally, the motion predictor models the conditional distribution :  
\begin{equation}
D(\theta) = p_{\theta}\big(\boldsymbol{S}^{t+1:T} \,\big|\, \boldsymbol{S}^{t-H:t}, M \big), \quad \boldsymbol{S}^{t+1:T} \sim D(\theta)
\label{motion_predictor}
\end{equation}
where the prediction depends on the past $H$ steps of agent states and the scenario map $M$. The predicted trajectory is typically represented as a sequence of future states drawn from a multivariate Gaussian distribution. Sampling from this distribution yields agents' states that define the next step in the evolving simulation scenario.

To capture the inherent uncertainty and multi-modality of future behaviors, we assume that the motion predictor outputs \( m \) possible future modes for each agent \( i \). The predictive distribution for agent \(i\) is thus represented as a weighted mixture of distributions: 
\begin{equation}
\label{baseline_predictor}
\resizebox{0.44\textwidth}{!}{$
\begin{aligned}
p_{\theta}\!\left(\boldsymbol{s}_i^{t+1:T} \,\middle|\, \boldsymbol{S}^{t-H:t}, M\right)
= \sum_{k=1}^m \pi_{i,k}\,
   p\!\left(\boldsymbol{s}_{i,k}^{t+1:T} \,\middle|\, \boldsymbol{S}^{t-H:t}, M\right)
\end{aligned}$}
\end{equation}
where \( \pi_{i,k} \) denotes the probability associated with the \(k\)-th predicted mode for agent \(i\). The learning objective of the motion predictor is to maximize the likelihood of the ground-truth trajectory by selecting the closest predicted mode and promoting a high probability for that mode.

The objective of risk scenario generation is to synthesize a sequence of agent states that may lead to AV evolving into collision or near-collision situations, given the recent driving context and map information. Recovering the true data distribution is inherently difficult due to the scarcity of real-world accident data. Considering the wide availability of large-scale datasets of normal human driving, we approximate the overall scene distribution as a mixture of nominal and risky regimes to efficiently exploit the limited risk data. Formally, the conditional generation process is defined as:
\begin{equation}
\resizebox{0.43\textwidth}{!}{$
\begin{aligned}
p\!\left(\mathbf{S}^{t+1:T} \,\middle|\, \mathbf{S}^{t-H:t}, M\right)  
&\approx (1 - \gamma_t)\, 
   p_{\theta}\!\left(\mathbf{S}^{t+1:T} \,\middle|\, \mathbf{S}^{t-H:t}, M\right)  \\
&\quad+\, \gamma_t\,
   p_{\theta^t_{\text{risk}}}\!\left(\mathbf{S}^{t+1:T} \,\middle|\, \mathbf{S}^{t-H:t}, M\right)
\end{aligned}$}
\label{eq:mix-prob}
\end{equation}
where $\gamma_t \in \{0,1\}$ is a risk indicator that controls the transition between nominal and risky dynamics based on the detected risk level at time $t$.
The nominal component is modeled by the motion predictor $D_\theta$ introduced in Eq.~(\ref{motion_predictor}). Meanwhile, the risk-aware distribution $p_{\theta^t_{\text{risk}}}$ is approximated through an optimization process that leverages crash data while retaining priors from normal driving behavior. Specifically, this process search for the optimal parameters $\Delta\theta$ such that $D_{\theta + \Delta\theta}$ better captures risky behavior within the VAE-learned latent space. Details of this procedure are provided in the following sections.

\begin{figure*}[!t]
\centering
\includegraphics[width=0.8\textwidth]{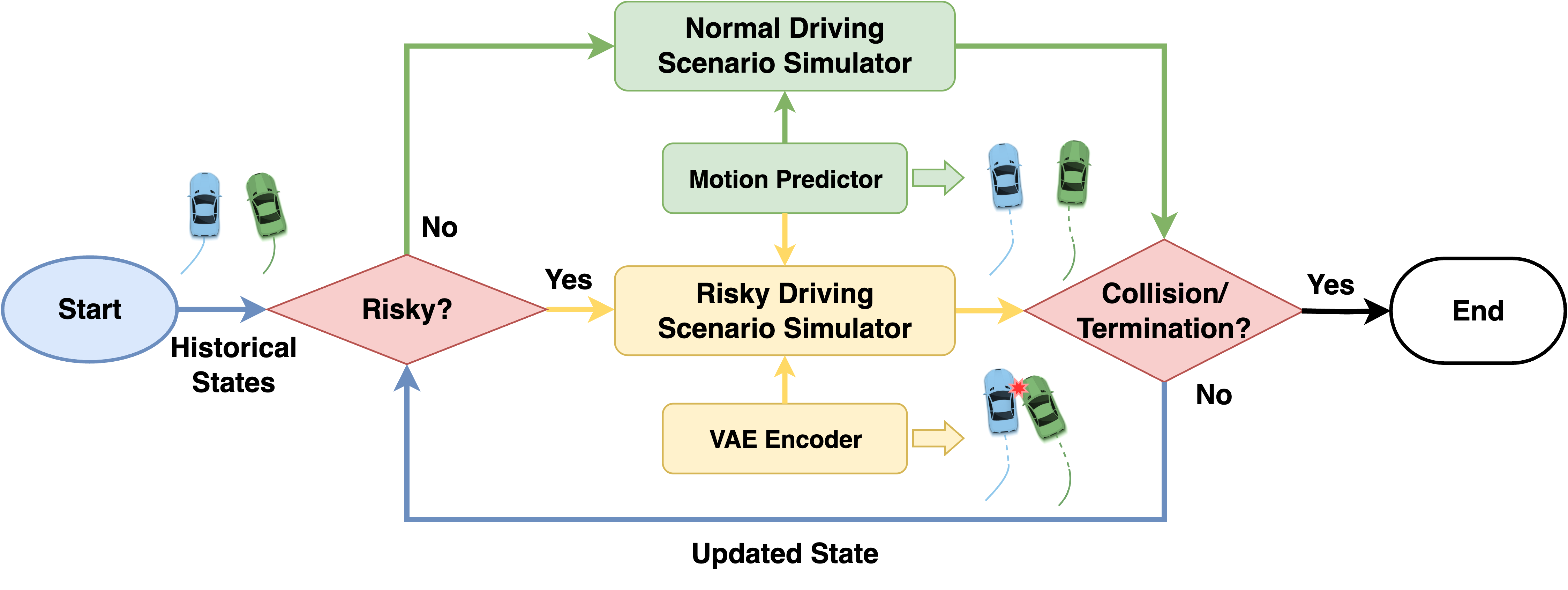}
\caption{Proposed risky testing scenario generation pipeline. }
\label{fig:sys_pip}
\end{figure*}
The closed-loop simulation process is illustrated in Fig.~\ref{fig:sys_pip}. The system first evaluates the current state to identify potential risk factors. In the absence of risks, the simulation continues with the normal driving scenario simulator, which is guided by the aforementioned motion predictor $D(\theta)$. Conversely, when risk factors are detected, the system activates the risky scenario simulator branch powered by the VAE encoder. The generated future states are subsequently evaluated by a collision state discriminator to decide whether the simulation should be terminated.This framework offers three key advantages: (i) it naturally integrates both normal and risky driving modes within a unified closed-loop process; (ii) it enables smooth transitions from nominal to risky conditions; and (iii) it ensures realistic scenario progression by terminating once a collision is detected.

Our method does not assume a specific architecture for the motion predictor given in Eq.~(\ref{motion_predictor}), provided that it supports backpropagation through all parameters $\theta$. Without loss of generality, we assume that the map can be vectorized into a feature representation and incorporated as part of the input. A representative sample architecture demonstrating this approach is described in the work of \cite{he2024knowmoformer}. By pretraining on large-scale open datasets, the motion predictor gains stronger generalization ability and is endowed with prior knowledge to better cope with unseen scenarios. In closed-loop testing, the AV operates independently, whereas the behaviors of background vehicles (BVs) are governed by the motion predictor. Importantly, the AV is modeled as a dummy agent within the predictor to facilitate realistic interactions.

\subsection{Risky Latent Space Construction}
\label{sub:risk_repre}
Differentiating risky from non-risky states in the original agent space is challenging, owing to data imbalance and the inherent diversity of risk cases. We mitigate this challenge by employing a VAE that learns latent representations of risk states in an unsupervised fashion, with training exclusively on risk data.

Variational Autoencoders (VAEs) are probabilistic generative models that encode input data into a latent distribution and decode samples from this distribution to reconstruct the input \cite{kingma2013auto}. Compared to Generative Adversarial Networks \cite{goodfellow2020generative}, VAEs offer stable, likelihood-based training that directly optimizes a variational objective. In contrast to diffusion models\cite{ho2020denoising}, which generate high-fidelity samples through iterative denoising, VAEs are more efficient to train and sample from. Crucially, VAEs are particularly well-suited for representation learning, as they encourage the discovery of structured and disentangled latent spaces \cite{higgins2017beta}.

In this work, VAE is adopted to efficiently capture the latent representation of original input features. The network consists of two neural networks: (i) an encoder \(q_\phi(z \mid \boldsymbol{f}_i)\), which maps an input feature \(\boldsymbol{f}_i\) to a distribution over latent variables \(z\); (ii) a decoder \(p_\gamma(\boldsymbol{f}_i \mid z)\), which reconstructs the input from the latent representation. The model is trained by maximizing the evidence lower bound (ELBO) with a standard normal prior \(p(z)=\mathcal{N}(0,I)\):

\begin{equation}
\label{eq:vae_loss}
\begin{aligned}
\mathcal{L}(\gamma,\phi; \boldsymbol{f}_i)
&= \mathbb{E}_{q_\phi(z \mid \boldsymbol{f}_i)}
      \!\left[\log p_\gamma(\boldsymbol{f}_i \mid z)\right]  \\
&\quad -\, 
      D_{\mathrm{KL}}\!\left(q_\phi(z \mid \boldsymbol{f}_i)\,\|\,p(z)\right)
\end{aligned}
\end{equation}
where the first term encourages faithful reconstruction of the input features \(\boldsymbol{f}_i\), and the second term regularizes the approximate posterior toward the prior. This framework facilitates the learning of a continuous and smooth latent representation that not only captures variations in the input space but also supports sampling and interpolation within the latent domain.

To improve learning efficiency under limited data availability, we introduce a feature preprocessing step designed to preserve essential relational structures. This preprocessing improves the model’s generalization ability to unseen states compared to training directly on raw state pairs.

\begin{figure}[!h]
  \centering
  \includegraphics[width=0.7\columnwidth]{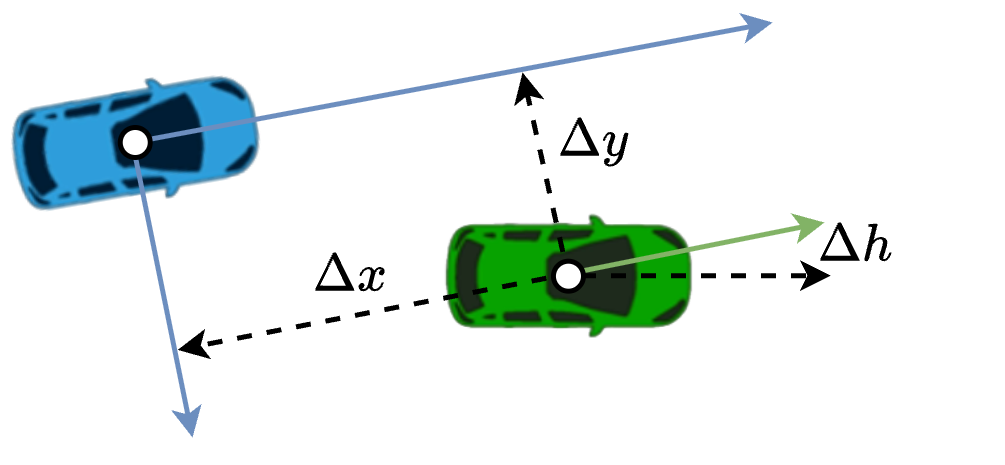}
  \caption{Agent poses represented in the AV coordinate system: AV (blue) and BV (green).}
  \label{fig:feature_rep}
\end{figure}

At each time step \(t\), we designate the AV as the ego vehicle and select a background vehicle BV$_i$ through a binary mask \(M_{a,i}\), which isolates their states from the scenario state \(\boldsymbol{S}^t\). Based on these states, a post-processed feature vector is constructed to capture their relative motion: 
\begin{equation}
\label{relative_input}
\resizebox{0.445\textwidth}{!}{$
\begin{aligned}
\boldsymbol{f}_i 
&= g\!\left(AV^t, \boldsymbol{s}_i^t\right) 
 = g\!\left(\boldsymbol{S}^t \odot M_{a,i}\right)                                      \\
&= \big[\,\Delta x^t,\;\Delta y^t,\;\cos(\Delta h^t),\;\sin(\Delta h^t),
          \;\Delta v^t,\;v_{\text{ego}}^t,\;v_i^t\,\big] \in \mathbb{R}^7
\end{aligned}$}
\end{equation}
where \(\Delta x\) and \(\Delta y\) denote the relative position of BV$_i$ in the AV's local coordinate frame, and \(\Delta h\) is the relative heading angle as illustrated in Fig.~\ref{fig:feature_rep}. In addition, \(\Delta v^t\) represents the relative speed (e.g., \(v_i^t - v_{\text{ego}}^t\)), while \(v_{\text{ego}}^t\) and \(v_i^t\) are the absolute speeds of the AV and BV$_i$, respectively.

We train the VAE exclusively on risky data using the input representation defined in Eq.~(\ref{relative_input}), enabling the encoder to learn latent features that capture the underlying structure of safety-critical behaviors and reveal dominant risk patterns in the data. Focusing the training on risky samples prevents bias toward nominal states and ensures that the learned latent space is specialized for risk characterization. The training process follows the standard VAE paradigm, where the encoder produces parameters \([\mu_\phi(\boldsymbol{f}_i),\,\sigma_\phi(\boldsymbol{f}_i)] = q_\phi(\boldsymbol{f}_i)\), from which latent variables are sampled via the reparameterization trick. The decoder then reconstructs the input to preserve essential relational features, guided by the training objective in Eq.~(\ref{eq:vae_loss}). After training, each pair \((\boldsymbol{S}^t, M_{a,i})\) is mapped to its corresponding latent distribution, providing a compact and continuous representation that supports sampling, interpolation, and downstream risk scenario generation. 

\begin{figure*}[!t]
  \centering
  \includegraphics[width=0.7\textwidth]{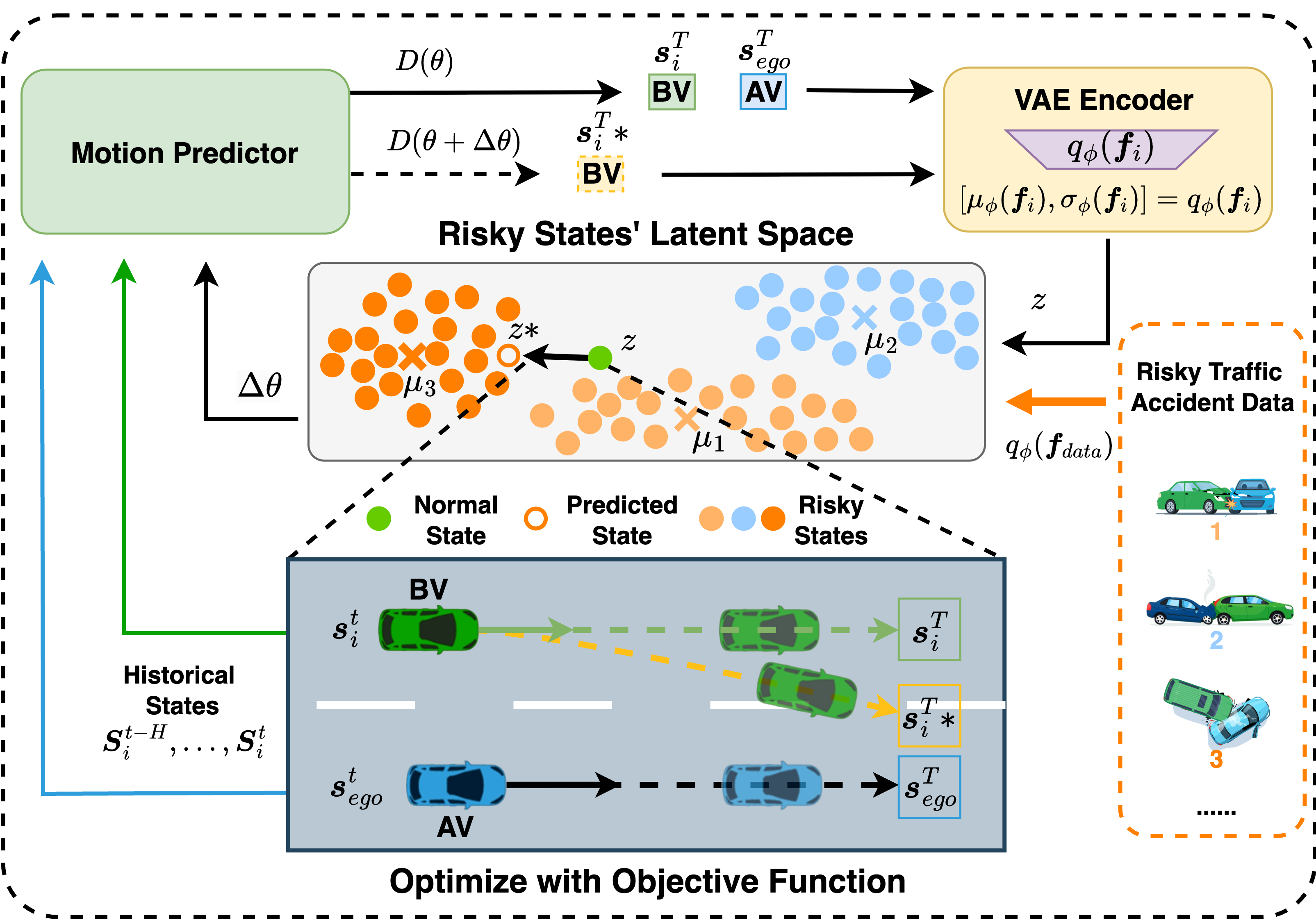}
  \caption{Illustration of the risk scenario generation framework. The system integrates a motion predictor, VAE-based latent representation, and latent-space optimization to produce dynamically plausible and controllable risky states.}
  \label{fig:sys_arch}
\end{figure*}

\subsection{Risky State Generation}
\label{sub:risky_state}
Unlike conventional trajectory-based formulations, which often require hand-crafted loss functions in the state space, we cast this task as a constrained optimization problem in the latent space of risky states:
\begin{equation}
\centering
 \begin{aligned}
\min_{\Delta \theta} \quad & \mathcal{L}( q_\phi(\boldsymbol{f}_i), p_{target}) \\
\textrm{s.t.} \quad & \boldsymbol{S}^{t+1:T} \sim D_{\theta + \Delta \theta}(\boldsymbol{S}^{t-H:t}), \\ 
&  \boldsymbol{f}_i = g(\boldsymbol{S}^T \odot M_{a, i}), \\
                    & \|\Delta \theta\| \leq \epsilon.
\end{aligned}
\label{opt:prob1}
\end{equation}
Here, $D_\theta$, introduced in Eq.~(\ref{motion_predictor}), denotes a pretrained motion predictor that produces initial predictions of the next $T-t$ steps given the past $H$ steps. The optimization searches for parameter perturbations $\Delta \theta$ within a bounded radius $\epsilon$, ensuring the resulting trajectories remain dynamically plausible. The feature vector $\boldsymbol{f}_i$ encodes the relative state between the AV and BV$_i$ at time $T$, which is mapped into the risky latent space by the VAE encoder. By minimizing the latent-space loss $\mathcal{L}$, the encoded representation is steered toward a target prior distribution $p_{target}$ that characterizes risky states. This formulation enables the generation of controllable risky scenarios, as illustrated in Fig.~\ref{fig:sys_arch}, by smoothly guiding predicted states toward regions associated with targeted high-risk outcomes. To further specify how this latent-space guidance is achieved during optimization, we next detail the construction of the target-specific objective.

\subsubsection{Design of Target-Specific Objective Function}
\label{subsub:obj}
In our framework, the concrete objective function for guiding a state toward a specific risky cluster $k$ is defined as  
\begin{equation}
\label{eq:loss}
\begin{aligned}
\mathcal{L}\!\left(q_\phi(\boldsymbol{f}_i),\, p_{\text{target}}\right)
&= \text{Dist}\!\left(q_\phi(\boldsymbol{f}_i),\, p_{\text{target}}\right)  \\
&\quad -\, w \cdot 
    \log\!\bigl(p(\mu_{\phi}(\boldsymbol{f}_i)\in C_k) + 10^{-8}\bigr)
\end{aligned}
\end{equation}
which jointly encourages high assignment probability and proximity in latent space to the target cluster. In this way, the updated motion predictor is guided to generate trajectories that are both classified into the correct risky category and remain close to prototypical risky configurations observed in the training data.  

To construct the target prior distribution, we formalize these prototypes by partitioning the latent space into $K$ clusters derived from the training data. Each cluster $C_i$ is approximated by a Gaussian distribution, with parameters estimated directly from the samples assigned to the cluster:
\begin{equation}
\mu_i = \frac{1}{|C_i|} \sum_{z \in C_i} z, 
\qquad
\sigma_i^2 = \frac{1}{|C_i|} \sum_{z \in C_i} \|z - \mu_i\|^2,
\end{equation}
leading to the cluster distribution:
\begin{equation}
p_{target}=p_{\text{coll},i}(z) = \mathcal{N}(z \mid \mu_i, \sigma_i^2 I), 
\quad i \in \{1, \dots, K\}.
\end{equation}

With the target prior distribution, we define the distance term $\text{Dist}(\cdot)$ between the encoded distribution $q_\phi(z \mid \boldsymbol{f}_i)$ and the cluster distribution $p_{\text{coll},k}(z)$ for a given state $\boldsymbol{f}_i$. For computational tractability, we approximate this distance by measuring the squared Euclidean distance between the mean of the encoded distribution, $\mu_\phi(\boldsymbol{f}_i)$, and the cluster mean $\mu_k$ as follows:  
\begin{equation}
\label{distance}
\text{Dist}(q_\phi(\boldsymbol{f}_i), p_{target}) = \tfrac{1}{2}\,\|\mu_\phi(\boldsymbol{f}_i) - \mu_k\|^2.
\end{equation}
Compared with Kullback–Leibler(KL) divergence, which requires evaluating full distributional differences and is often sensitive to variance mismatch, this distance-based formulation provides both computational efficiency and intuitive interpretability. 

Furthermore, to estimate the likelihood of a state belonging to each cluster, we employ a distance-based Softmax function, as shown in Eq.~(\ref{likelihood}),  which assigns higher probabilities to clusters that are closer in the latent space: 
\begin{equation}
\label{likelihood}
\begin{aligned}
p\!\left(\mu_{\phi}(\boldsymbol{f}_i)\in C_k\right)
&= 
\frac{\exp\!\left[-\text{Dist}\!\left(q_\phi(\boldsymbol{f}_i),\, p_{\mathrm{coll},k}\right)\right]}
     {\sum_{j=1}^{K} \exp\!\left[-\text{Dist}\!\left(q_\phi(\boldsymbol{f}_i),\, p_{\mathrm{coll},j}\right)\right]}  \\[4pt]
&= 
\frac{\exp\!\left[-\tfrac{1}{2}\,\bigl\|\mu_\phi(\boldsymbol{f}_i)-\mu_k\bigr\|^2\right]}
     {\sum_{j=1}^{K} \exp\!\left[-\tfrac{1}{2}\,\bigl\|\mu_\phi(\boldsymbol{f}_i)-\mu_j\bigr\|^2\right]} .
\end{aligned}
\end{equation}

\subsubsection{Optimization for Risky States}
As illustrated in Section~\ref{subsub:obj}, Specifically, the sampled future state $S^T \odot M_{a,i}$ is passed through the trained VAE encoder to compute $\mathcal{L}(q_\phi(\boldsymbol{f}_i), p_{\text{coll},k})$. During the risky motion simulation phase, the optimization problem defined in Eq.~(\ref{opt:prob1}) is solved by backpropagation. The VAE parameters remain fixed, while the gradient of the objective function is propagated through the motion predictor $D(\theta)$. The predictor parameters $\theta$ are iteratively updated, steering the generated trajectories toward the targeted risky distribution in the latent space.
\begin{algorithm}[!h]
\caption{Risk State Optimization via Latent-Space Guidance}
\label{alg:risk_opt}
\textbf{Input: } History states $\boldsymbol{S}^{t-H:t}$; motion predictor $D(\theta)$; 
trained VAE encoder $q_\phi$; target prior $p_{\text{coll},k}$; 
risk threshold $\tau$; step size schedule $\{\eta_r\}$; max steps $R$.\\
\textbf{Output: }Generated risk trajectory $\boldsymbol{S}^{t+1:T}$. 
\vspace{0.5em} \\ 
\textbf{Risk detection at step $t$:}\; \\
Sample (or roll out) a provisional future 
$\boldsymbol{S}^{t+1:T} \sim D_{\theta}(\boldsymbol{S}^{t-H:t})$.\; \\ 
Construct the candidate BV set $\mathcal{I}_t$ based on their relative distance to the AV.\;
\vspace{0.5em} \\ 
\textbf{for}{ $i \in \mathcal{I}_t$ }\textbf{do} \\ 
\hspace*{1.5em} Form the pair $(AV, i)$ and compute $\boldsymbol{f}_i = g(\boldsymbol{S}^T \odot M_{a,i})$.\; \\ 
\hspace*{1.5em} Evaluate the risk-oriented loss $\mathcal{L}(q_\phi(\boldsymbol{f}_i),\, p_{\text{coll},k})$.\;
\vspace{0.5em} \\ 
\textbf{if}{ $\min_{i \in \mathcal{I}_t}\mathcal{L}(q_\phi(\boldsymbol{f}_i), p_{\text{coll},k}) > \tau$ }\textbf{then}
\vspace{0.2em} \\ 
\hspace*{1.5em} \textbf{return }$(\theta, \boldsymbol{S}^{t+1:T})$ \\
\textbf{else} \\
\hspace*{1.5em} Targeted BV $\leftarrow \arg\min_{i \in \mathcal{I}_t} \mathcal{L}(q_\phi(\boldsymbol{f}_i), p_{\text{coll},k})$ \\
\vspace{0.5em} \\ 
\textbf{Optimization loop:} Set $\theta_0 \leftarrow \theta$.\; \\ 
\textbf{for }{$r = 0$} \textbf{to }{$R-1$}\textbf{ do}\\ 
\hspace*{1.5em} Roll out $\boldsymbol{S}^{t+1:T} \sim D_{\theta_r}(\boldsymbol{S}^{t-H:t})$.\;\\
\hspace*{1.5em} Evaluate objective $\mathcal{L}(q_\phi(\boldsymbol{f}_i), p_{\text{coll},k})$.\;\\
\hspace*{1.5em} Backpropagate to obtain $u_r \leftarrow \nabla_{\theta}\mathcal{L}(\theta_r)$ (VAE frozen).\;\\
\hspace*{1.5em} Update $\tilde{\theta}_{r+1} \leftarrow \theta_r - \eta_r u_r$.\;\\
Generate final $\boldsymbol{S}^{t+1:T} \sim D_{\theta_R}(\boldsymbol{S}^{t-H:t})$ and \textbf{return} $(\theta_R, \boldsymbol{S}^{t+1:T})$.
\end{algorithm}

Let \(u_r\) denote the update direction at step \(r\) and \(\eta_r\) the step size. The update rule is given by
\begin{equation}
\theta_{r+1} \,=\, \theta_r \,-\, \eta_r\,u_r,
\end{equation}
which after \(R\) steps leads to  
\begin{equation}
\resizebox{0.43\textwidth}{!}{$
\theta_R = \theta_0 - \sum_{r=0}^{R-1} \eta_r u_r
\Rightarrow
\|\theta_R - \theta_0\|_2 = \Bigl\| \sum_{r=0}^{R-1} \eta_r u_r \Bigr\|_2.$}
\end{equation}
By the triangle inequality, we obtain 
\begin{equation}
\|\theta_R - \theta_0\|_2 \;\le\; \sum_{r=0}^{R-1} \eta_r\,\|u_r\|_2.
\end{equation}
Assuming a constant learning rate \(\eta_r=\eta\) and uniformly bounded update norms \(\|u_r\|_2 \le U_{\max}\), this simplifies to 
\begin{equation}\label{eq:upper_bd}
\|\theta_R - \theta_0\|_2 \;\le\; R\,\eta\,U_{\max}.
\end{equation}

Thus, the number of steps $R$ and the learning rate ${\eta}$ can be jointly selected to ensure that the parameter updates remain within the perturbation upper bound $\epsilon$ specified in Eq.~(\ref{opt:prob1}). For clarity, the optimization process is summarized in Algorithm~\ref{alg:risk_opt}. Alternative objective functions $\mathcal{L}$ can be employed, provided they effectively capture the mismatch between the encoded distribution and the prior. The effect of different distance metrics is examined in the ablation study presented in Section~\ref{subsec:abla}.

\section{Experimental Setup}
\label{sec:setup}
This section presents the experimental setup used to evaluate the proposed framework. We first describe the preparation of crash datasets corresponding to the target scenarios. Next, we introduce the baseline algorithms adopted for comparison and outline the evaluation metrics employed to assess safety-critical relevance. Finally, we provide implementation details, including network architectures, training configurations, and simulation parameters used throughout all experiments.

\subsection{Synthetic Collision Dataset}
We focus on two representative scenarios: one-way traffic and an intersection. The one-way traffic scenario emphasizes longitudinal interactions and lateral maneuvers within structured lane configurations, enabling targeted analysis of car-following dynamics and lane-change maneuvers. In contract, the intersection scenario features multidirectional traffic flows and heterogeneous road users, including vulnerable road users (VRUs) such as pedestrians and cyclists. These elements give rise to complex conflict points and a broader spectrum of accident typologies. Together, these two scenarios capture both structured and unstructured interaction patterns, providing a comprehensive basis for evaluating the effectiveness of the proposed method in generating diverse safety-critical conditions for AVs. 

Since the primary objective is to generate challenging cases for AVs, we focus on common crash categories in which the AV is highly likely to be at fault. For the one-way traffic scenario, we synthesize three categories of crashes for the AV: front-impact, left sideswipes, and right sideswipes. To this end, temporally desynchronized vehicle trajectories extracted from the highD dataset~\cite{krajewski2018highd} were employed. The generation process begins by identifying lane-changing vehicles from the dataset and designating them as ego vehicles. At the completion of a lane-change maneuver, the ego's state is used as a reference to search for surrounding vehicles within its local vicinity, irrespective of frame alignment. Candidate BVs are selected based on spatial proximity—within a radial distance and relative kinematic conditions. Specifically, for front-impact collisions, the lead vehicle is required to have a lower longitudinal velocity than the ego. These constraints ensure that the synthesized collisions are both physically plausible and representative of common accident typologies. The generated crashes are categorized and validated using the proposed collision classification method described in Section~\ref{subsec:eval_metrics}. Following this procedure, 7{,}000 risky states are generated for each of the three categories.

For the intersection scenario, crashes are generated by executing the data-driven motion predictor \cite{lin2025intersectionde} in a closed-loop configuration until collisions occurred. Owing to the presence of heterogeneous road users and multidirectional flows, the resulting accidents included diverse subtypes such as vehicle–vehicle and vehicle–cyclist collisions. Consistent with the one-way traffic scenario, collision types are determined from the ego vehicle's perspective and mapped into three classes: front-impact, left sideswipe, and right sideswipe. And again 7{,}000 risky states are collected for each collision type. 

\subsection{Baseline Algorithm}
\begin{figure}[!h]
  \centering
  \includegraphics[width=0.5\columnwidth]{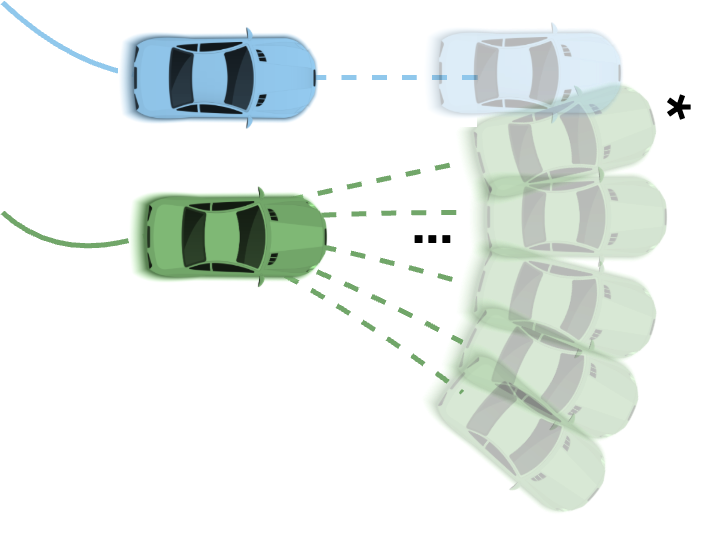}
  \caption{Generation of a risky scenario using the baseline algorithm, where the mode with the highest risk level is selected.}
  \label{fig:base_line}
\end{figure}

Because the motion predictor is primarily trained on nominal driving data, it rarely produces meaningful accident cases when used directly in closed-loop simulation. To establish a reference point for evaluation, we define a baseline algorithm that explicitly enforces the selection of collision-prone modes. At each simulation step, the algorithm selects one background vehicle together with one of its candidate modes that is most likely to yield a collision with the AV’s most likely mode, as illustrated in Fig.\ref{fig:base_line}. This design provides a deterministic yet risk-oriented alternative against which the proposed CRAG can be compared.

Concretely, for each BV \(i\) and its \(j\)-th candidate mode, we construct the post-processed feature vector \(\boldsymbol{f}_{i,j}\) using the AV's most likely mode combined with the \(j\)-th mode of vehicle \(i\). The encoded distribution of this feature vector, denoted by \(q_\phi(\boldsymbol{f}_{i,j})\), is then evaluated using the distance-based risk surrogate \(\mathcal{L}(\cdot,\cdot)\) defined in Section~\ref{subsub:obj}. The baseline algorithm selects the pair ($i^*$, $j^*$) that minimizes the distance to the target risky distribution:
\begin{equation}
\label{eq:baseline}
(i^*, j^*) \;=\; \arg\min_{i,j} \; \mathcal{L}\!\big(q_\phi(\boldsymbol{f}_{i,j}),\,p_{coll, k}\big),
\end{equation}
the selected background vehicle $i^*$ is then rolled out using its chosen mode \(j^*\), while all remaining agents are simulated using their most likely modes.

This baseline therefore serves as a risk-maximizing heuristic: it systematically biases the simulation toward producing collisions but does not incorporate gradual risk transitions or the broader context of traffic interactions. Consequently, it provides a useful reference for evaluating the enhanced realism and controllability introduced by CRAG.

\subsection{Evaluation Metrics}
\label{subsec:eval_metrics}
To evaluate the performance of the simulated agents, we assess both \emph{controllability} and \emph{collision efficiency} using two complementary metrics. The \emph{general collision rate} (GCR) measures the overall frequency of collisions and is defined as the fraction of simulation runs that result in at least one collision:
\begin{equation}
\mathrm{GCR} = \frac{N_{\mathrm{coll}}}{N_{\mathrm{total}}},
\end{equation}
where $N_{\mathrm{coll}}$ denotes the number of runs with detected collisions and $N_{\mathrm{total}}$ is the total number of runs.  

The \emph{targeted collision rate} (TCR) further refines this measure by considering only collisions of the designated type:
\begin{equation}
\mathrm{TCR} = \frac{N_{\mathrm{target}}}{N_{\mathrm{total}}},
\end{equation}
where $N_{\mathrm{target}}$ is the number of runs that result in a collision of the specified type. Together, GCR and TCR provide a comprehensive assessment of both the overall collision likelihood and the controllability of generating specific risky scenarios.

Furthermore, to assess the realism of the generated collisions, we measure the distributional differences of $\Delta x$ and $\Delta y$  relative to the collected crash data. A higher KL divergence indicates a greater deviation from the training distribution used to construct the latent space.
To determine collision types, we combine the \emph{Separating Axis Theorem} (SAT) with the \emph{Minimum Translation Vector} (MTV). Each vehicle $i$ is modeled as a rigid rectangle centered at $\boldsymbol{c}_i = (x_i, y_i)$ with heading angle $\theta_i$. Its longitudinal and lateral unit vectors are given by $\boldsymbol{e}_i^x = \bigl(\cos \theta_i,\; \sin \theta_i \bigr),  \boldsymbol{e}_i^y = \bigl(-\sin \theta_i,\; \cos \theta_i \bigr).$
For an ego vehicle (AV) and a background vehicle $i$ (BV), the displacement vector is $\boldsymbol{d}_i = \boldsymbol{c}_i - \boldsymbol{c}_{ego}$. Collision detection is carried out by projecting $\boldsymbol{d}_i$ onto the candidate axes of the ego vehicle, $\boldsymbol{u} \in \{\boldsymbol{e}_{ego}^x, \boldsymbol{e}_{ego}^y\}$. The projection radius of vehicle $i$ along axis $\boldsymbol{u}$ is
\begin{equation}
    \label{eq:r}
r_i = a \, \big|\boldsymbol{e}_i^x \cdot \boldsymbol{u}\big| + b \, \big|\boldsymbol{e}_i^y \cdot \boldsymbol{u}\big|,
\end{equation}
where $a=L/2$ and $b=W/2$ denote half-length and half-width of the vehicle, respectively.  

\begin{figure}[!h]
  \centering
  \includegraphics[width=0.95\columnwidth]{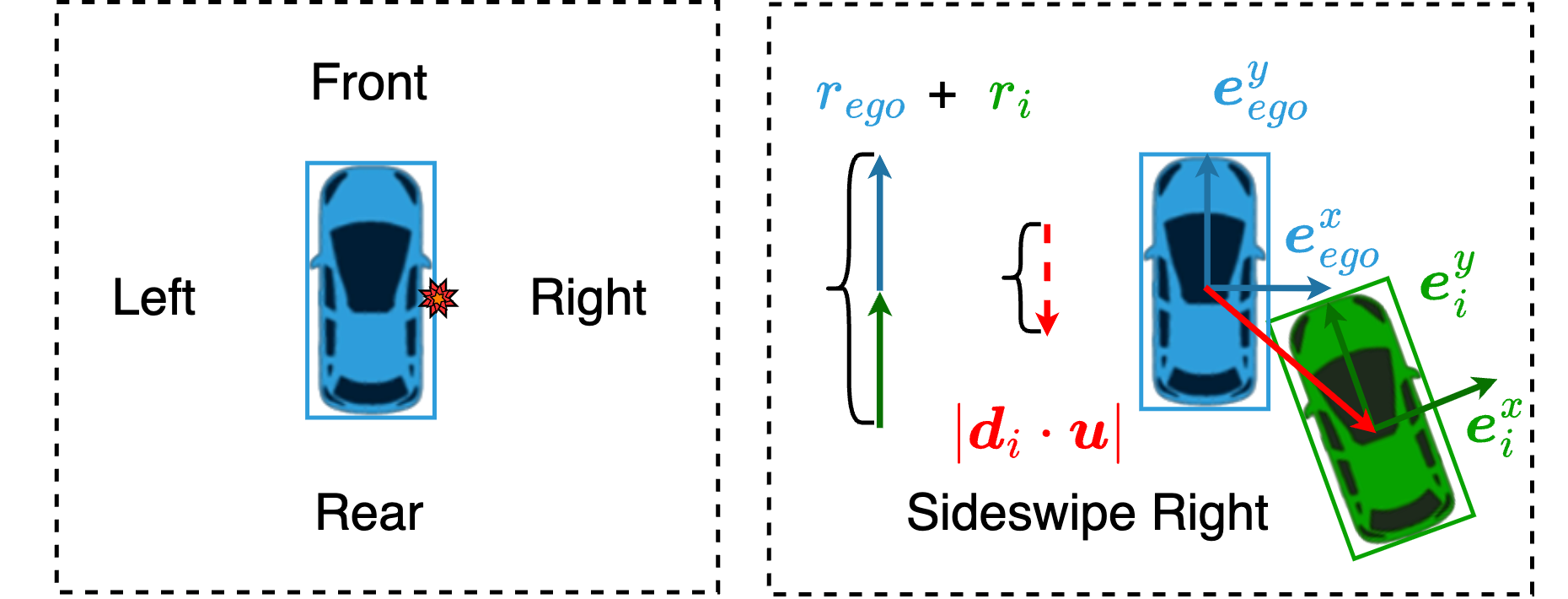}
  \caption{Determination of collision type based on penetration depth and MTV.}
  \label{fig:col_detect}
\end{figure}

The penetration depth along axis $\boldsymbol{u}$ is then defined as
\begin{equation}
\delta_u = (r_1 + r_2) - \big|\boldsymbol{d}_i \cdot \boldsymbol{u}\big|,
\end{equation}
where $r_1$ and $r_2$ are the projection radii of the two vehicles given by Eq.~(\ref{eq:r}). The MTV is defined as the smallest displacement vector that, when applied to one vehicle, resolves the overlap. Its direction corresponds to the axis $\boldsymbol{u}$ with the minimum $\delta_u$, and its magnitude equals $\delta_u$.   If the projected intervals are disjoint on any axis, the vehicles are non-colliding; otherwise, an overlap exists, indicated by $\delta_u > 0$. Based on the direction and magnitude of $\delta_u$, collisions are classified into four categories: front, rear, left sideswipe, and right sideswipe, as illustrated in Fig.~\ref{fig:col_detect}. This procedure ensures consistent and physically interpretable categorization of collision outcomes.

\subsection{Implementation Details}
We employ a lightweight MLP-based VAE to encode relative-motion features \(\boldsymbol{f}_i \in \mathbb{R}^7\) into a low-dimensional latent space \(\mathbb{R}^{5}\) and reconstruct them back. The encoder \(q_\phi(z\mid \boldsymbol{f}_i)\) consists of two fully connected layers with 256 and 128 hidden units, respectively, each followed by ReLU activations. Batch normalization is applied to the input features, and a dropout rate of 0.1 is applied after the second hidden layer. Two linear heads parameterize the mean and log-variance of a diagonal Gaussian posterior, with \(\log\sigma_\phi^2(\boldsymbol{f}_i)\) predicted for numerical stability. The decoder \(p_\gamma(\boldsymbol{f}_i\mid z)\) mirrors the encoder with two fully connected layers (128 and 256 units, ReLU) and a linear projection to \(\mathbb{R}^{7}\). Dropout (0.1) is applied before the final projection, and the reconstructed output is batch-normalized to stabilize training across heterogeneous feature scales. This design yields an efficient yet expressive latent representation of risky states.
Two different motion predictors are employed to validate the effectiveness of the proposed framework. For the one-way traffic scenario, we adopt a knowledge-conditioned Transformer as the motion predictor~\cite{he2024knowmoformer}, which captures structured lane interactions and generates realistic vehicle behaviors. For the intersection scenario, we employ a scene-aware Transformer-based motion predictor that incorporates traffic light control and jointly models both vehicles and vulnerable road users, enabling rich interaction simulation within the intersection.
All experiments were conducted on an NVIDIA RTX 3090 GPU. The optimization procedure uses the ADAM optimizer with a fixed learning rate \(\eta\), applied over \(T\) update steps. For typical hyperparameter settings (\(T=30\), \(\eta=10^{-5}\)), the numerical bound $ \|\theta_T - \theta_0\|_2 \;\le\; 3\times10^{-4}\,U_{\max},$ can be obtained, as derived in Eq.~(\ref{eq:upper_bd}). By constraining the updated parameters remain within a controlled range relative to \(\theta_0\), the procedure ensures both optimization stability and compliance with the constraints in Eq.~(\ref{opt:prob1}).

\section{Results and Analysis}
\label{sec:exp}
In this section, we first analyze the learned latent space structure and verify the interpretability of the clustering results. Next, we evaluate the performance of the proposed method across targeted driving scenarios to demonstrate its capability in generating targeted safety-critical situations. Finally, we conduct ablation studies to assess the contribution and effectiveness of the core modules.

\subsection{Embedding Space Analysis}
\begin{figure*}[!t]
\centering
\includegraphics[width=0.42\textwidth]{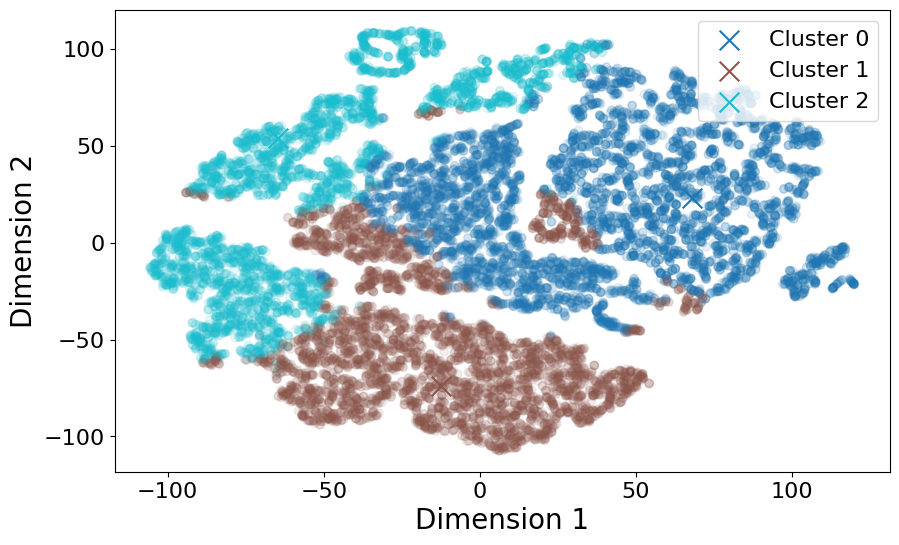}
\includegraphics[width=0.42\textwidth]{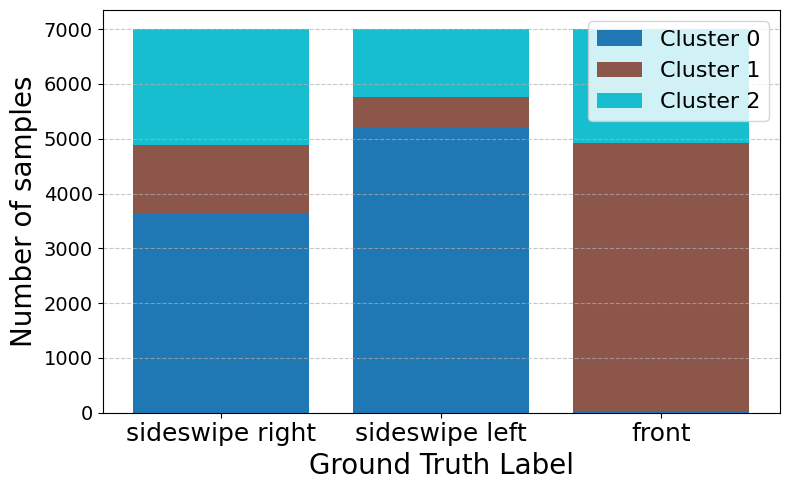}
\includegraphics[width=0.42\textwidth]{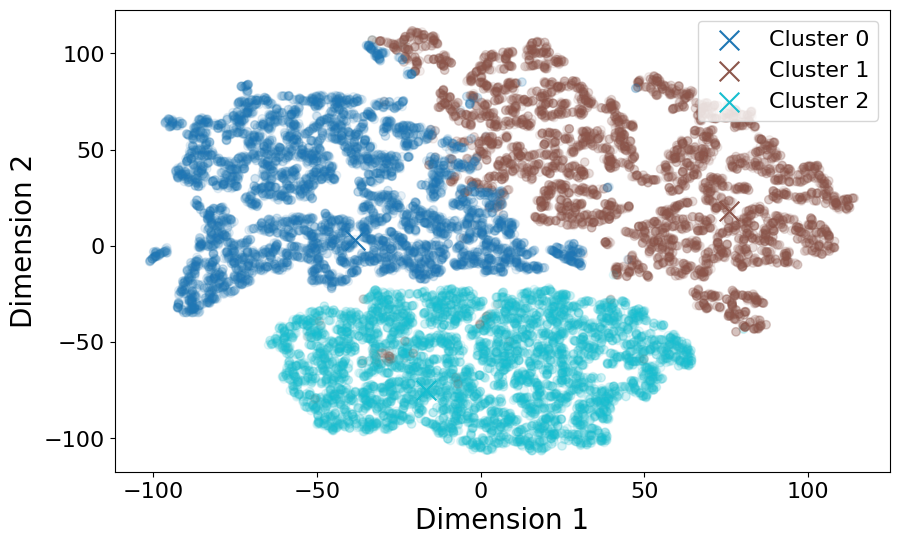}
\includegraphics[width=0.42\textwidth]{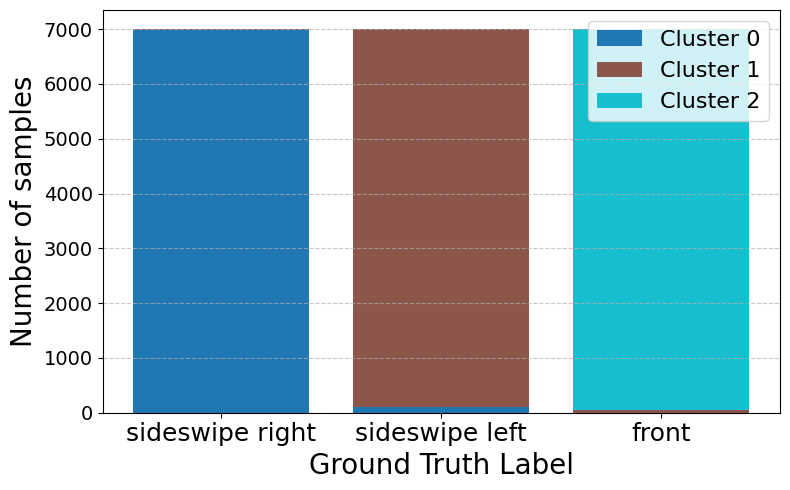}
\caption{Comparison of clustering results in the raw input feature space (top) and the VAE-learned latent space (bottom). Left panels show t-SNE visualizations with K-means assignments; right panels show the distribution of accident categories within each cluster. }
\label{fig:vae}
\end{figure*}

A structured latent space supports precise sampling and enables controllable generation of targeted risk scenarios. We compare clustering performance on raw input features against that within the learned latent space for one-way traffic scenario.  As shown in Fig.~\ref{fig:vae}, the clusters in the learned latent space exhibit higher purity, indicating that the learned representations are not only well separated but also strongly aligned with their semantic categories. In contrast, clustering in the original input space fails to achieve a clear separation among the three collision types. Quantitatively, clustering in the latent space attains a purity of 0.99, significantly outperforming raw feature clustering (purity 0.58). These results underscore that the VAE-learned latent space effectively encodes discriminative features, enabling more reliable identification of distinct collision states.

\begin{figure*}[!h]
\centering
\includegraphics[width=0.7\textwidth]{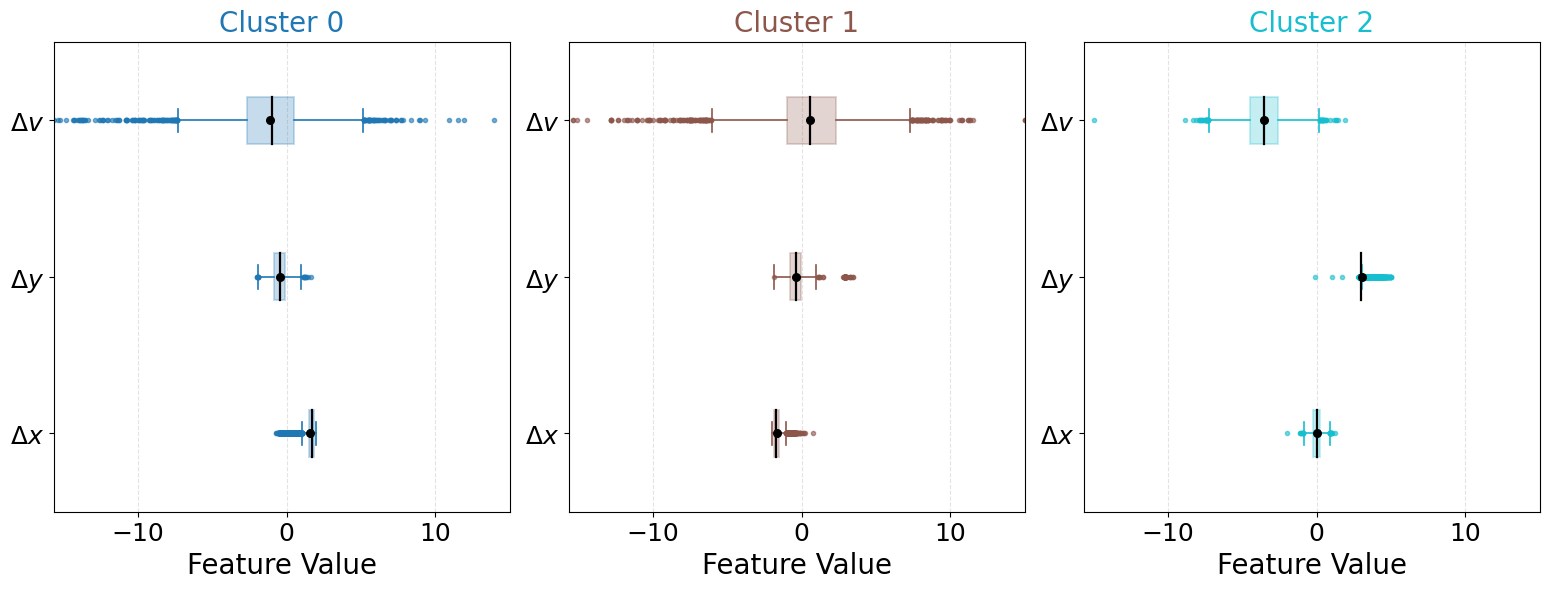}
\caption{Box plots of feature distributions across clusters,  visualized using features that significantly distinguish different collision types. }
\label{fig:vae_box}
\end{figure*}

As shown in Fig.~\ref{fig:vae_box}, the box plots reveal distinctive feature patterns for each cluster. For cluster 0, which is dominated by right-swipe collisions of the AV, the relative displacement $\Delta x$ is predominantly positive and emerges as the most discriminating feature. Cluster 1, dominated by left-swipe collisions, instead shows negative values of $\Delta x$. In contrast, cluster 2, characterized by AV collisions with vehicles directly in front, is primarily associated with positive values of $\Delta y$. These results confirm that the latent space not only separates risky states but also aligns with intuitive physical interpretations of vehicle interactions during accidents.

\begin{figure*}
\centering
\includegraphics[width=0.7\textwidth]{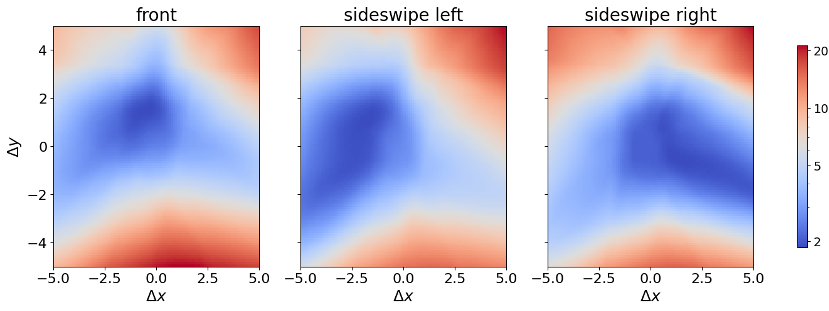}
\caption{Visualization of distance landscapes within the learned latent space. The heatmaps illustrate how the distance from encoded states to their respective cluster centers (computed using Eq.~(\ref{distance})) varies as the input features $\Delta x$ and $\Delta y$ change, with other features fixed at their mean values.} \label{fig:vae_mse_map}
\end{figure*}

To further investigate the structure of the learned latent space, we vary the input features $\Delta x$ and $\Delta y$ fed into the VAE encoder while keeping the other features fixed at their average values. We then compute the distances of the resulting embeddings to their respective cluster centers in the latent space. As shown in Fig.~\ref{fig:vae_mse_map}, the minimal distance regions align well with the patterns observed in the box plots: for instance, the left-swipe cluster shows minimal distance when $\Delta x$ is positive, while other collision types exhibit corresponding consistency. The resulting heatmaps also demonstrate that the latent space is smooth and responds continuously to variations in the input features. This behavior suggests that the latent representation is well organized, mapping semantically similar states to nearby positions. Consequently, searching for risky states in the latent space can be approximated by exploring variations in the original input features. 

\begin{figure*}[!ht]
  \centering
\includegraphics[width=0.7\textwidth]{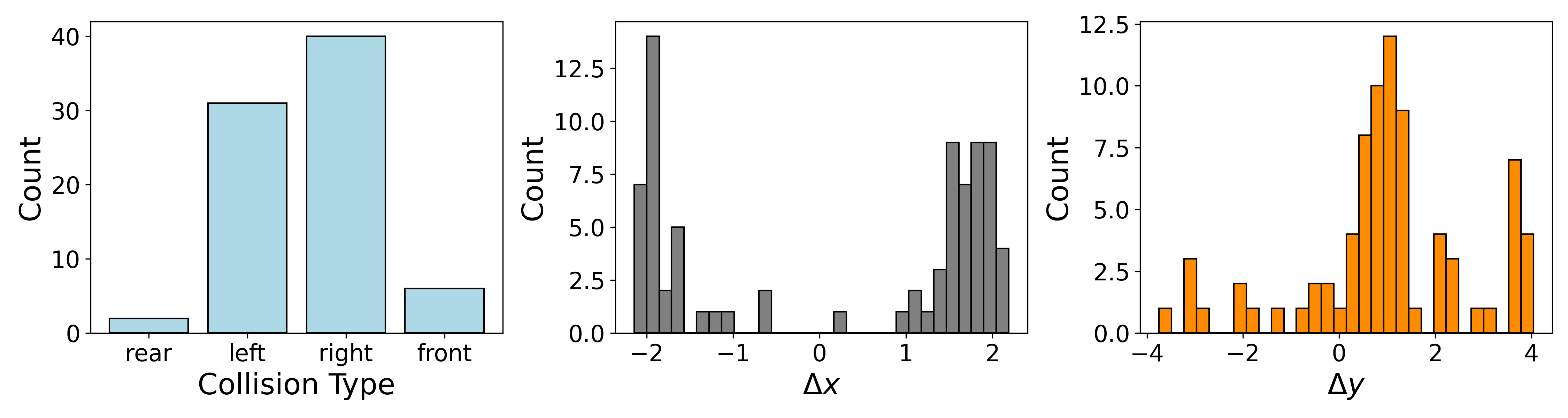}
\includegraphics[width=0.7\textwidth]{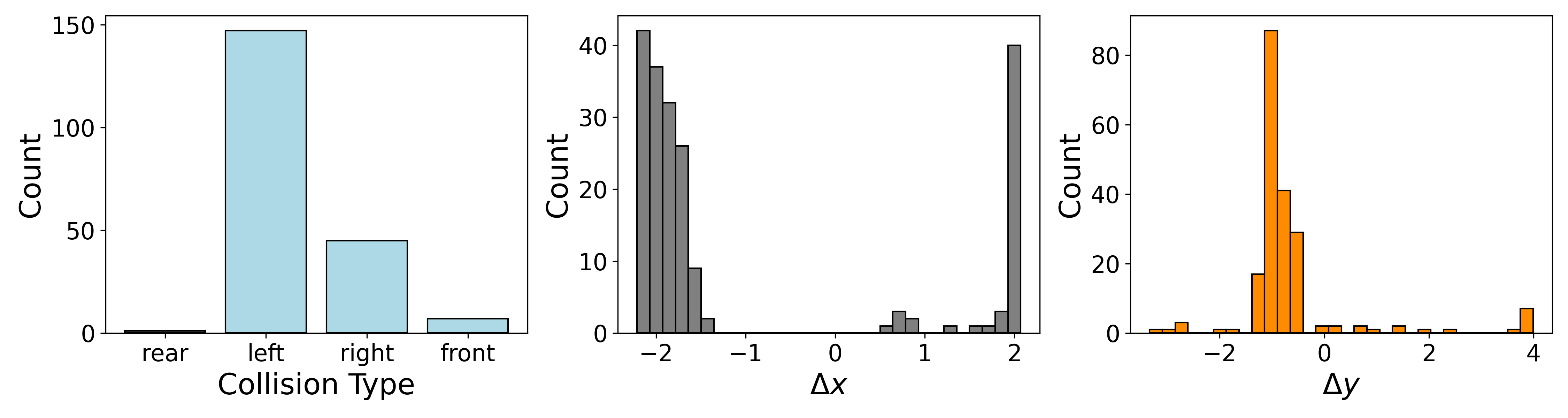}
\includegraphics[width=0.7\textwidth]{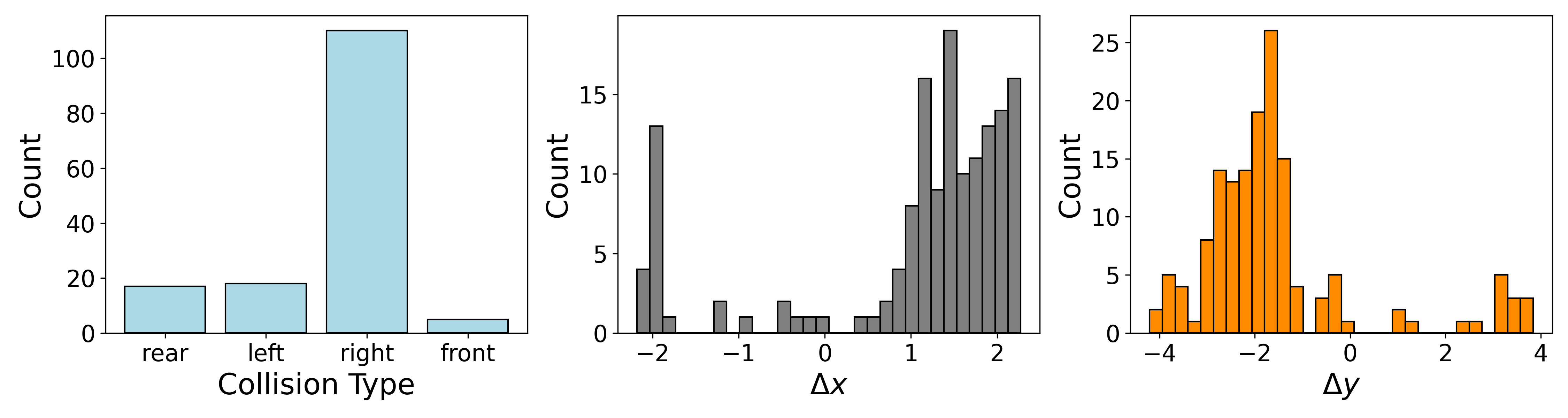}
  \caption{Distribution of generated collisions for targeted types (front, left, and right; shown from top to bottom). Left: distribution of generated collision types. Middle: distribution of final-state $\Delta x$. Right: distribution of final-state $\Delta y$. }
  \label{fig:gen_col_dist_one_way}
\end{figure*}

In summary, the latent space learned by VAE provides a compact and semantically meaningful representation that separates collision types more effectively than raw features. This property enables not only robust identification of risky states but also provides a stable basis for controlled risk scenario generation, which we further investigate in the following section.
\subsection{Risk Scenario Generation}
To comprehensively assess the effectiveness and generality of CRAG, we investigate the capability of CRAG to generate diverse and safety-critical situations through a three-stage analysis: first examining its behavior in structured one-way traffic, then validating its robustness in more intricate intersection environments, and finally conducting an ablation study to disentangle the roles of individual design components.
\subsubsection{One-Way Traffic Scenario}
We first evaluate the quantitative performance of CRAG against the baseline algorithm. To ensure statistical reliability, we conduct 1,000 simulation runs for both the baseline and CRAG, with initial states sampled from the highD dataset to remain consistent with real-world driving distributions. As reported in Table~\ref{tbl1:collision}, the baseline achieves relatively low collision rates (1.7 \%, 2.9 \%, 4.5 \%), whereas our algorithm substantially increases collision likelihoods (20 \%, 15.0 \%, 7.9 \%), enabling significantly more efficient testing. This outcome is expected: the baseline predictor preserves only the six most likely modes from training, leaving rare but safety-critical cases underrepresented. By contrast, CRAG introduces explicit guidance in the latent space, steering the motion predictor toward novel and risky combinations that would otherwise remain unexplored. The distinction is particularly evident for sideswipes, where CRAG achieves targeted collision rates of 14.7 \% and 11.0 \%, compared to only 1.2 \% and 0.7 \% under the baseline. 
\begin{table*}[!ht]
\centering
\caption{Comparison of collision rates and KL divergences between the baseline and CRAG in the one-way scenario. Results are reported for General Collision Rate (GCR), Targeted Collision Rate (TCR), and KL divergences of feature distributions ($D_{KL}(\Delta x)$ and $D_{KL}(\Delta y)$). Arrows indicate whether higher ($\uparrow$) or lower ($\downarrow$) values are preferred.}
\label{tbl1:collision}
\begin{tabular*}{\textwidth}{@{\extracolsep{\fill}}lcccccc}
\toprule
& \multicolumn{2}{c}{Left sideswipe} 
& \multicolumn{2}{c}{Right sideswipe} 
& \multicolumn{2}{c}{Front-impact} \\
\cmidrule(lr){2-3} \cmidrule(lr){4-5} \cmidrule(lr){6-7}
Metric 
& Baseline & CRAG 
& Baseline & CRAG 
& Baseline & CRAG \\
\midrule
GCR (\%)$\uparrow$          & 1.7 & \textbf{20.0} & 2.9 & 15.0 & 4.5 & 7.9 \\
TCR (\%)$\uparrow$          & 1.2 & \textbf{14.7} & 0.7 & 11.0 & 1.6 & 0.6 \\
$D_{KL}(\Delta x)\downarrow$ & 1.1 & 0.7           & 1.1 & \textbf{0.3} & 15.0 & 0.3 \\
$D_{KL}(\Delta y)\downarrow$ & 17.4 & 0.5          & 4.2 & 3.4          & \textbf{0.1} & 0.3 \\
\bottomrule
\end{tabular*}
\end{table*}

\begin{table*}[!ht]
\centering
\caption{Comparison of collision rates and KL divergences between the baseline and CRAG in the intersection scenario. Results are reported for General Collision Rate (GCR), Targeted Collision Rate (TCR), and KL divergences of feature distributions ($D_{KL}(\Delta x)$ and $D_{KL}(\Delta y)$). Arrows indicate whether higher ($\uparrow$) or lower ($\downarrow$) values are preferred.}
\label{tbl3:collision_inter}
\begin{tabular*}{\textwidth}{@{\extracolsep{\fill}}lcccccc}
\toprule
& \multicolumn{2}{c}{Left sideswipe} & \multicolumn{2}{c}{Right sideswipe} & \multicolumn{2}{c}{front-impact} \\
\cmidrule(lr){2-3}\cmidrule(lr){4-5}\cmidrule(lr){6-7}
Metric & Baseline & CRAG & Baseline & CRAG & Baseline & CRAG \\
\midrule
GCR (\%)$\uparrow$ & 5.2 & 16.0 & 4.6 & 11.3 & 5.6 & \textbf{19.3} \\
TCR (\%)$\uparrow$ & 1.0 & 5.7 & 1.4 & \textbf{5.9} & 1.2 & 3.5 \\
$D_{KL}(\Delta x)$$\downarrow$  & 9.0 & 1.1 & 8.5 & 1.0 & 0.4 & \textbf{0.3} \\
$D_{KL}(\Delta y)$$\downarrow$  & 5.0 & 0.3 & 5.0 & \textbf{0.2} & 4.9 & 2.0 \\
\bottomrule
\end{tabular*}
\end{table*}

The generated distributions in Fig.~\ref{fig:gen_col_dist_one_way} further validate these findings. For front-impact collisions, the histogram of $\Delta y$ shows a strong bias toward positive values, indicating that the BV is positioned ahead of the AV, which is consistent with a front-bumper impact from the ego. In contrast, left-sideswipe and right-sideswipe cases exhibit distinctive $\Delta x$ distributions: left-sideswipes are characterized by negative $\Delta x$, while right-sideswipes correspond to positive $\Delta x$. These clear directional signatures align with the semantic definitions of the collision types and confirm that CRAG does not merely inflate collision counts, but generates states with feature distributions consistent with realistic accident patterns. Moreover, the spread of $\Delta y$ values within sideswipe cases illustrates that collisions occur under varied longitudinal offsets, highlighting the diversity of generated terminal crash configurations. Furthermore, compared to the feature distributions of targeted collision cases generated by the baseline algorithm, CRAG consistently achieves lower  $D_{KL}$ across features, as shown in Table~\ref{tbl1:collision}. 

\begin{figure}
  \centering
\includegraphics[width=0.5\textwidth]{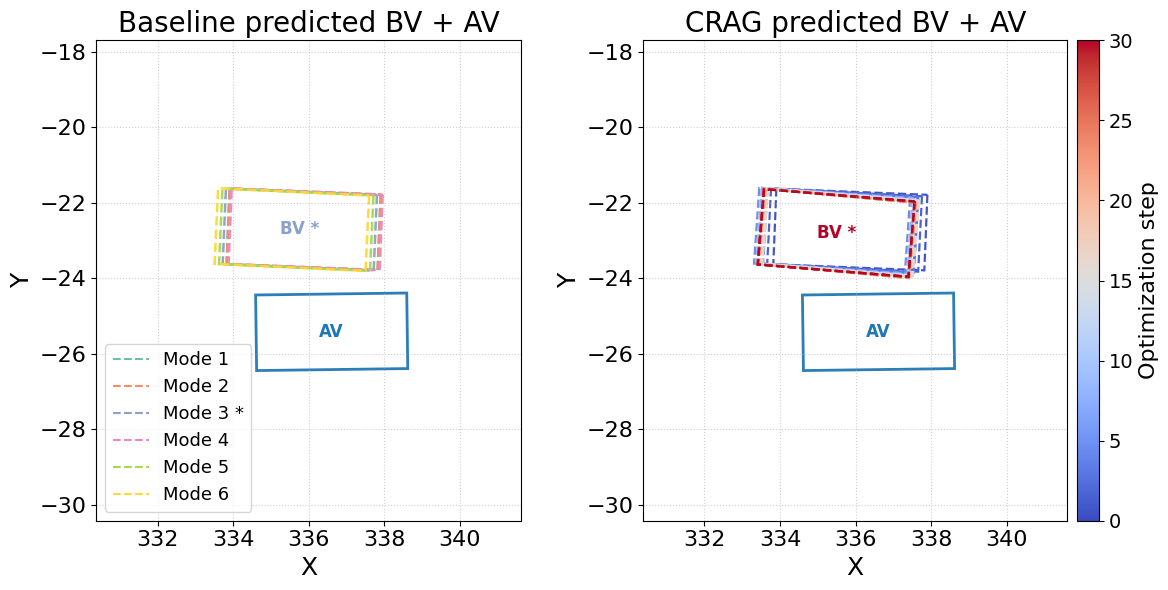}
  \caption{Qualitative comparison of CRAG and the baseline algorithm in  single-step prediction, with the best result marked by *. }
  \label{fig:opt_vis}
\end{figure}
\begin{figure*}
  \centering
  \begin{subfloat}{(a)}
    \centering
    \includegraphics[width=\textwidth]{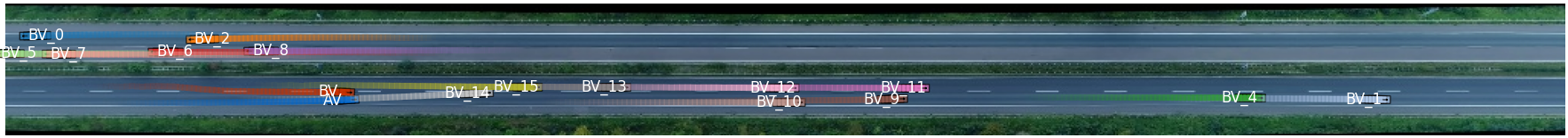}
  \end{subfloat}
  \begin{subfloat}{(b)}
    \centering
    \includegraphics[width=\textwidth]{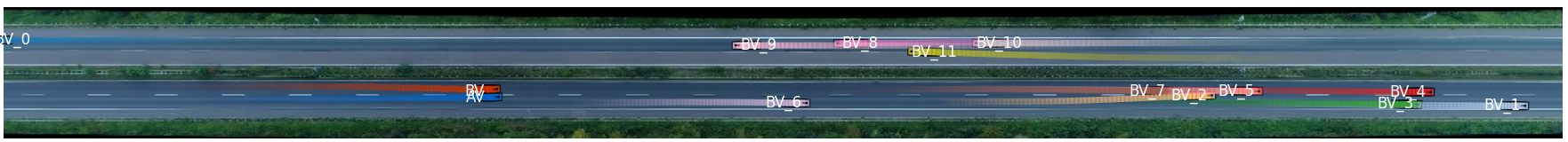}
  \end{subfloat}
  \begin{subfloat}{(c)}
    \centering
    \includegraphics[width=\textwidth]{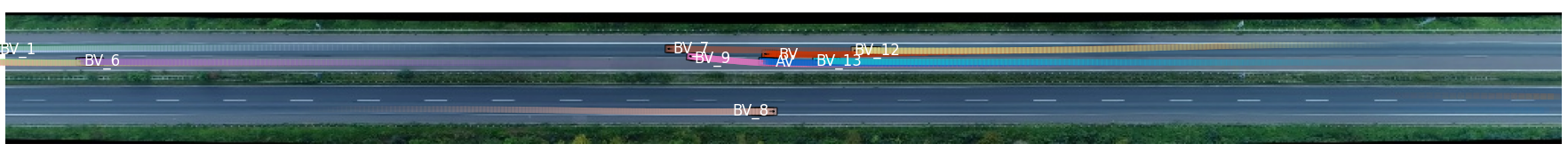}
  \end{subfloat}
  \begin{subfloat}{(d)}
    \centering
    \includegraphics[width=0.998\textwidth]{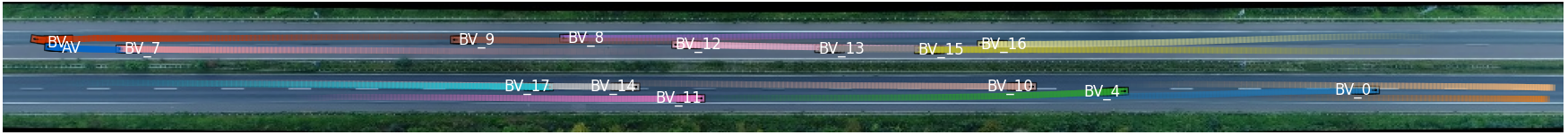}
  \end{subfloat}
  \begin{subfloat}{(e)}
    \centering
    \includegraphics[width=\textwidth]{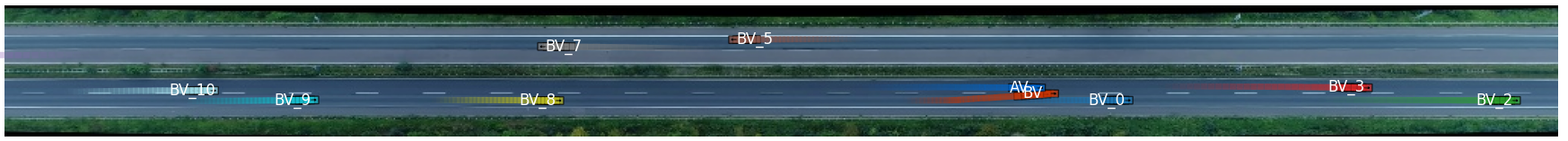}
  \end{subfloat}
  
 \caption{Sampled generations with collisions (AV in blue, BV in red). (a)–(b) left sideswipes; (c)–(d) right sideswipes; (e) front collision.}
 \label{fig:gen_col_eg}
\end{figure*}

To provide qualitative insights, Fig.~\ref{fig:opt_vis} illustrates how the CRAG algorithm and the baseline generate different future trajectories for the same historical states. Under CRAG, the identified BV gradually approaches the AV over 30 optimization steps, ultimately resulting in a left-sideswipe configuration. In contrast, the baseline modes mainly capture longitudinal speed variations but lack the directional diversity needed to represent right-lane change behaviors. Further examples are shown in Fig.~\ref{fig:gen_col_eg} to illustrate the diversity of generated collisions: (a) opposing lane changes by the AV and BV lead to a left-sideswipe; (b) the AV merges left with insufficient headway, resulting in another left-sideswipe; (c) the AV follows a lead vehicle into a right-lane change while a BV closes from behind, causing a right-sideswipe; (d) concurrent rightward merges by the AV and BV lead to insufficient clearance and a right-sideswipe; and (e) a BV cuts in from the left, where the AV fails to decelerate in time, producing a front collision. These cases highlight diverse causal factors—including inadequate gap acceptance, concurrent merges, and cut-ins—that give rise to complex multi-agent interactions.

\begin{figure*}
  \centering
\includegraphics[width=1.0\textwidth]{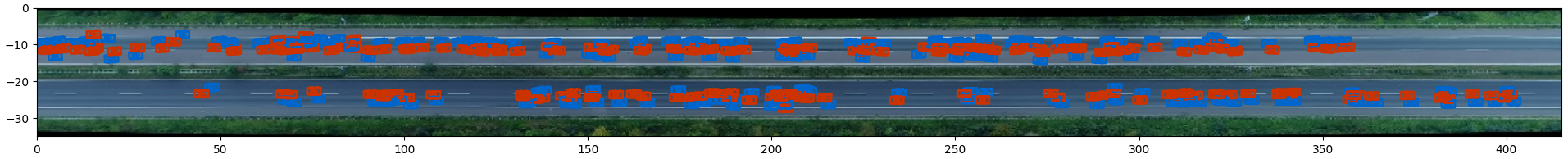}
  \caption{Diverse final collision states across map locations for left-sideswipe cases (AV in blue, BV in red). }
  \label{fig:left_collide_loc}
\end{figure*}

Finally, as shown in Fig.~\ref{fig:left_collide_loc}, the generated collisions occur at diverse map locations and exhibit substantial variability in terminal crash states for left-sideswipe cases. This demonstrates that CRAG not only increases collision frequency but also expands the coverage of risky scenarios in both spatial distribution and causal mechanisms. To the best of our knowledge, no prior work has achieved controllable induction of specified accident types. Our method fills this gap by systematically analyzing the accident data space and guiding a nominal predictor toward desired risk characteristics, thereby enabling scalable stress testing of AV systems.

\subsubsection{Intersection Scenario}
Intersections are inherently more complex than one-way traffic scenarios due to the presence of heterogeneous road users and the convergence of multidirectional traffic flows. VRUs are particularly safety-critical, as their behaviors are less predictable and require heightened attention from autonomous vehicles. Compared to vehicles in structured lane-following traffic, VRUs exhibit greater behavioral variability, engage in denser and more uncertain interactions. These characteristics significantly increase the difficulty of decision-making for AV systems, underscoring the necessity of intersection-specific risk scenario testing.

Quantitatively, as reported in Table~\ref{tbl3:collision_inter}, CRAG achieves consistently higher collision rates than the baseline across all three collision categories. In particular, targeted collision rates (TCR) increase by a substantial margin: CRAG yields $\mathbf{5.7\%}$, $\mathbf{5.9\%}$, and $\mathbf{3.5\%}$ for left sideswipes, right sideswipes, and frontal impacts, respectively, compared to baseline values of $1.0\%$, $1.4\%$, and $1.2\%$. At the same time, KL divergence values of the final-state feature distributions ($\Delta x, \Delta y$) are 
greatly reduced under CRAG, suggesting that the generated risky states more closely align with the prototypical risk clusters observed in training data. This confirms that CRAG not only improves controllability toward specific accident types but also preserves fidelity to real-world risky behaviors.

\begin{figure*}
  \centering
\includegraphics[width=0.72\textwidth]{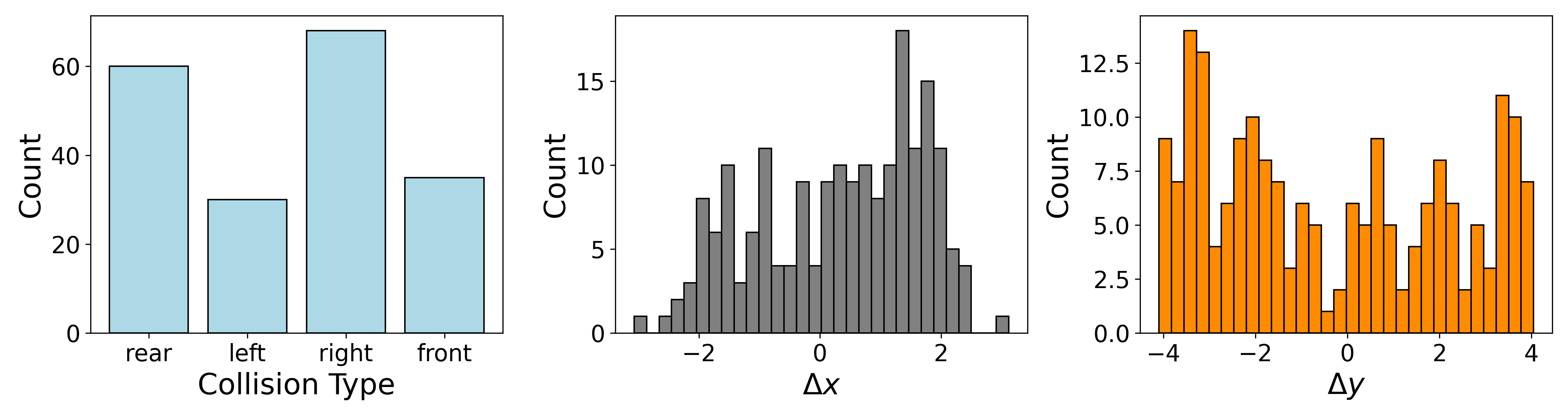}
\includegraphics[width=0.72\textwidth]{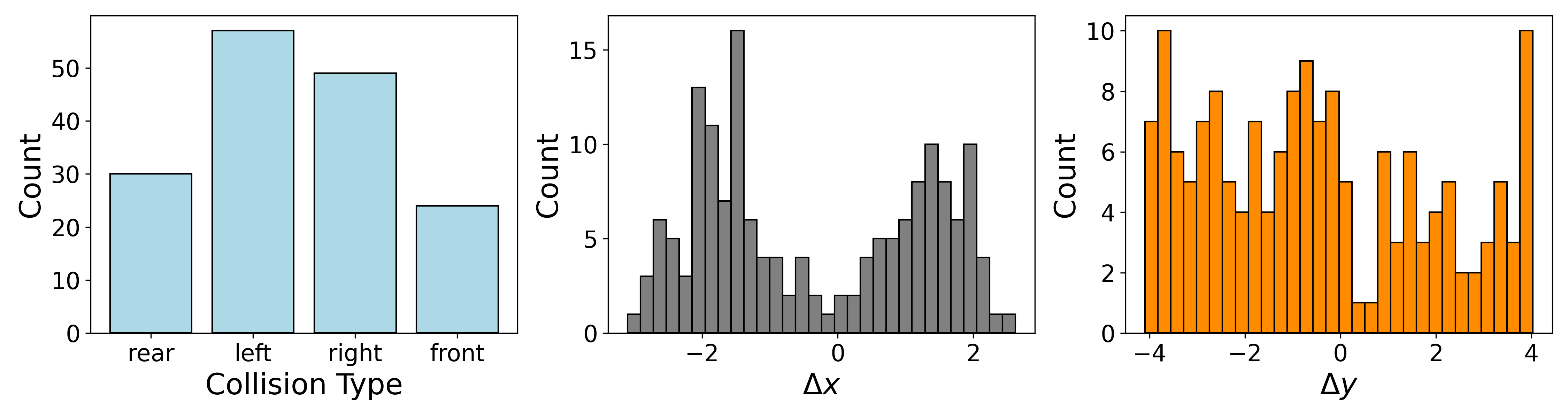}
\includegraphics[width=0.72\textwidth]{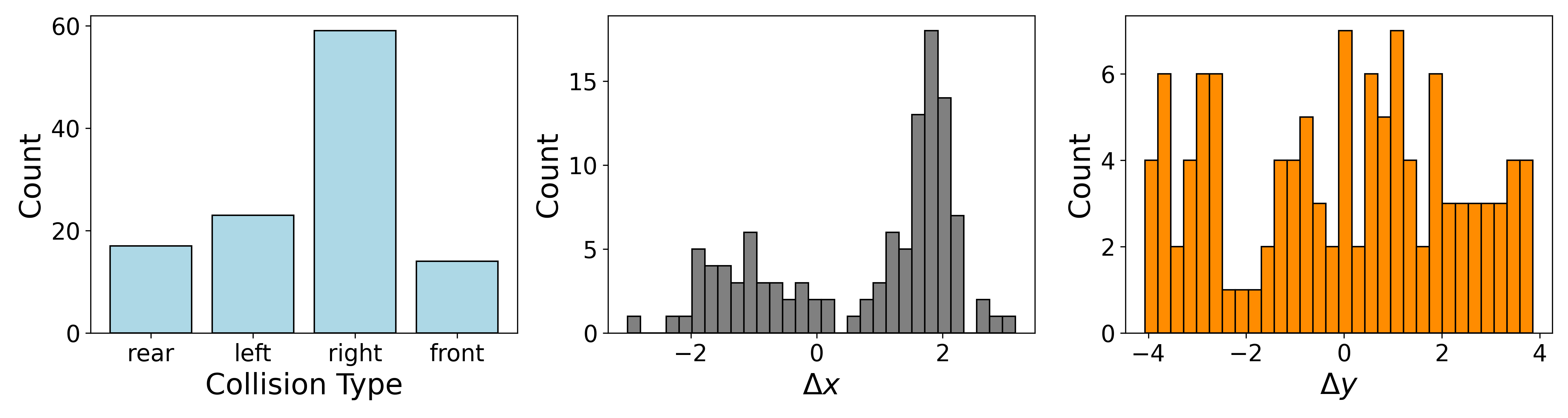}
  \caption{Distribution of generated collisions for targeted types (front, left, and right; shown from top to bottom). Left: distribution of generated collision types. Middle: distribution of final-state $\Delta x$. Right: distribution of final-state $\Delta y$. }
  \label{fig:gen_col_dist_intersection}
\end{figure*}

\begin{table*}[!h]
\caption{Ablation study on the design of the targeting loss. 
Results are reported for General Collision Rate (GCR), Targeted Collision Rate (TCR), and KL divergences of feature distributions ($D_{KL}(\Delta x)$ and $D_{KL}(\Delta y)$). 
Arrows indicate whether higher ($\uparrow$) or lower ($\downarrow$) values are preferred.}
\label{tb:ab}
\begin{tabular*}{\textwidth}{@{\extracolsep{\fill}}lcccc}
    \toprule
    Method & GCR (\%) $\uparrow$ & TCR (\%) $\uparrow$ & $D_{KL}(\Delta x)$ $\downarrow$ & $D_{KL}(\Delta y)$ $\downarrow$ \\
    \midrule
    Ours (distance-softmax + distance) & 15.0 & \textbf{11.0} & \textbf{0.3} & \textbf{3.4} \\
    MSE loss (mean-only)               & \textbf{19.9} &  9.7 & 0.5 & 5.6 \\
    Wasserstein loss                   & 17.7 &  8.5 & 0.6 & 5.5 \\
    KL divergence loss                 &  0.0 &  0.0 &  -  &  -  \\
    \bottomrule
\end{tabular*}
\end{table*}

Fig.~\ref{fig:gen_col_dist_intersection} presents the distributional characteristics of generated collisions for each targeted type. The left column shows the category distribution of collision cases generated by CRAG, where the targeted collision type consistently ranks among the most frequent categories. The middle and right columns display the final-state displacement distributions.

Left sideswipes are characterized by dominant negative $\Delta x$ values, while right sideswipes concentrate on positive $\Delta x$ values, consistent with their physical definitions. Front collisions show clear positive $\Delta y$, reflecting that the BV is typically located ahead of the AV at impact. Compared with the one-way traffic scenario, the distributions here exhibit broader spread and heavier tails, reflecting the larger variability in approach angles and maneuver strategies at intersections. This broader coverage demonstrates CRAG’s ability to generate diverse terminal crash states that capture the inherent uncertainty of real-world urban intersections.

\begin{figure*}
  \centering
\fbox{\includegraphics[width=0.42\textwidth]{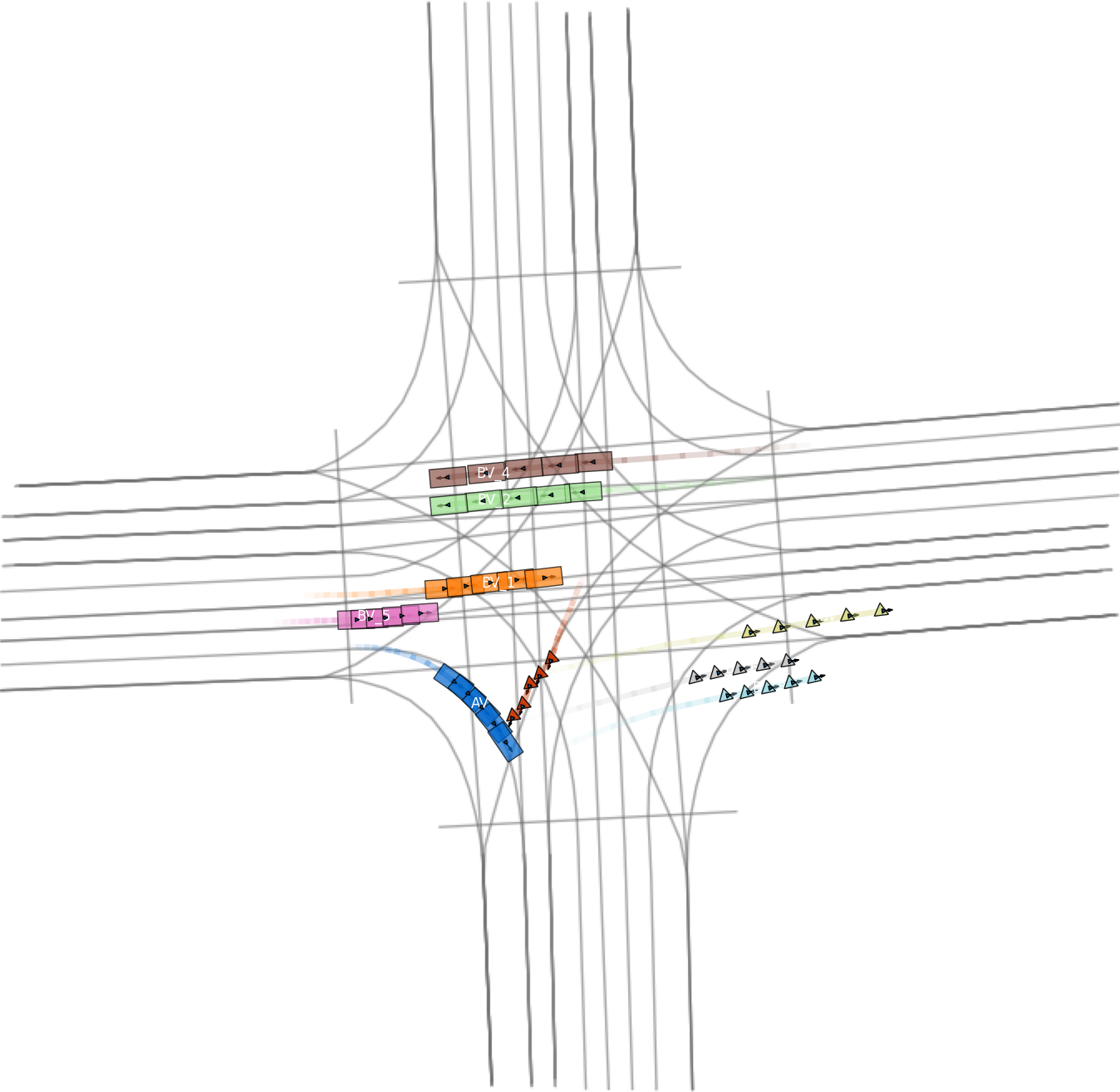}}
\fbox{\includegraphics[width=0.42\textwidth]{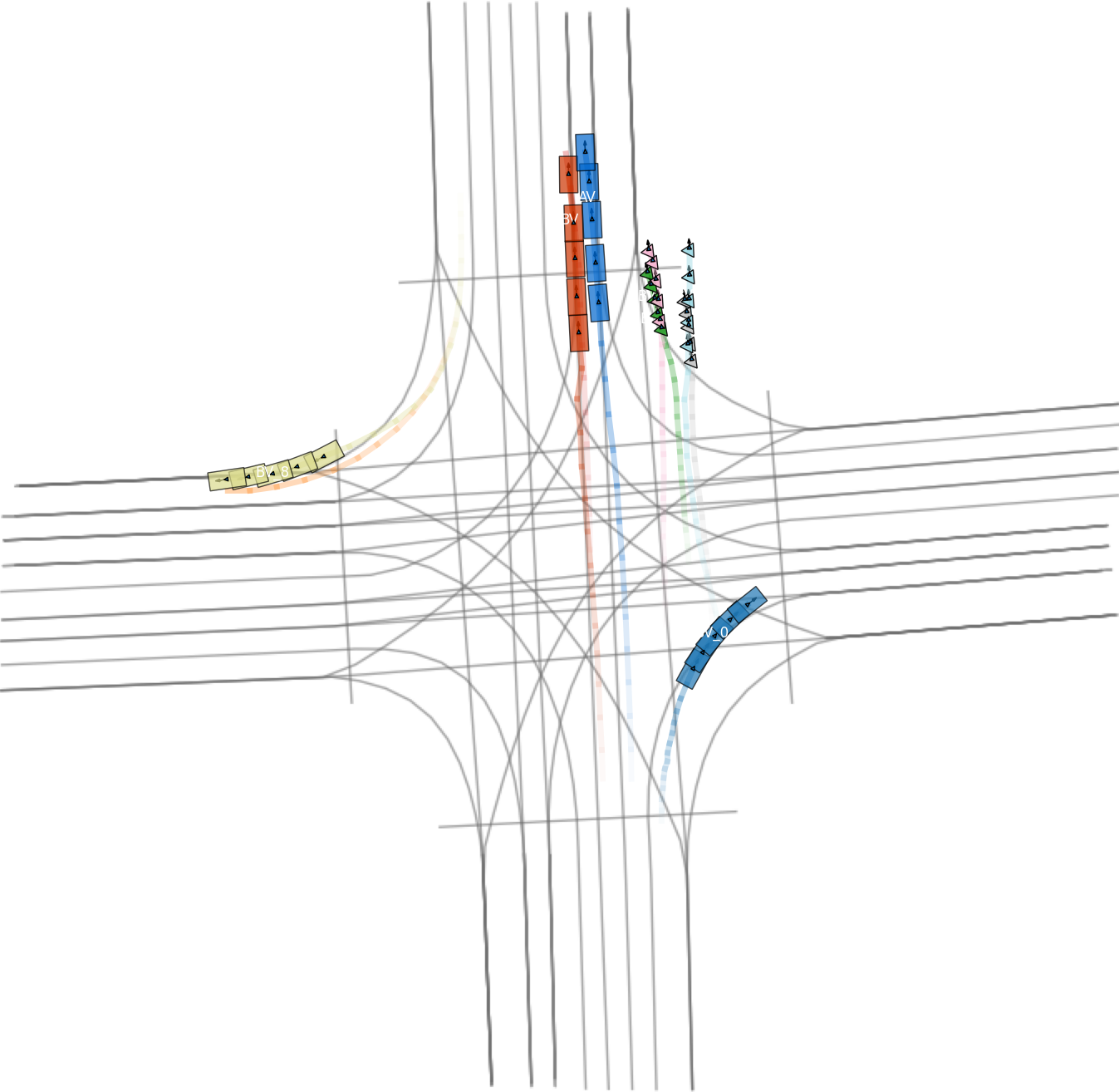}}
\fbox{\includegraphics[width=0.42\textwidth]{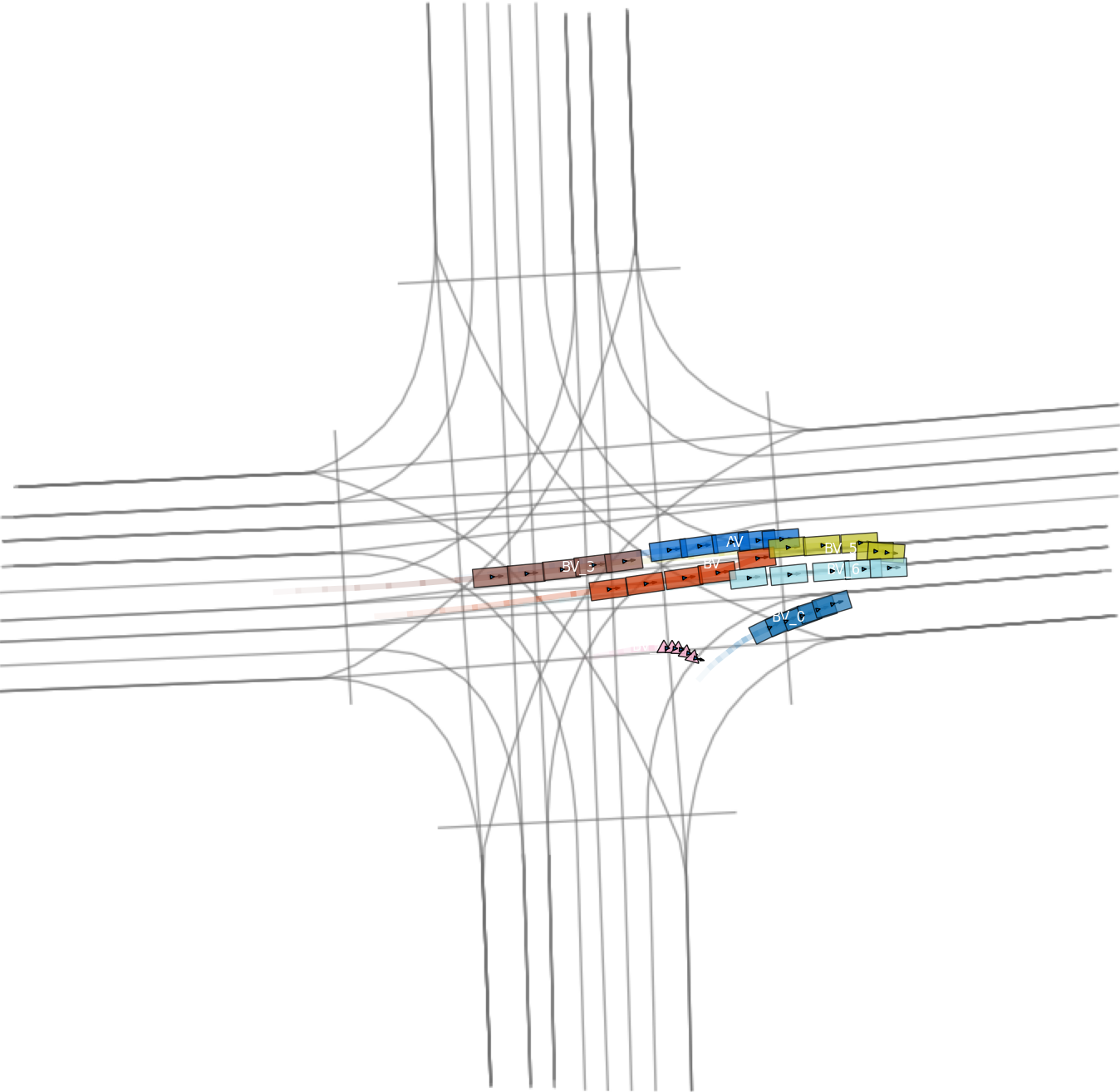}}
\fbox{\includegraphics[width=0.42\textwidth]{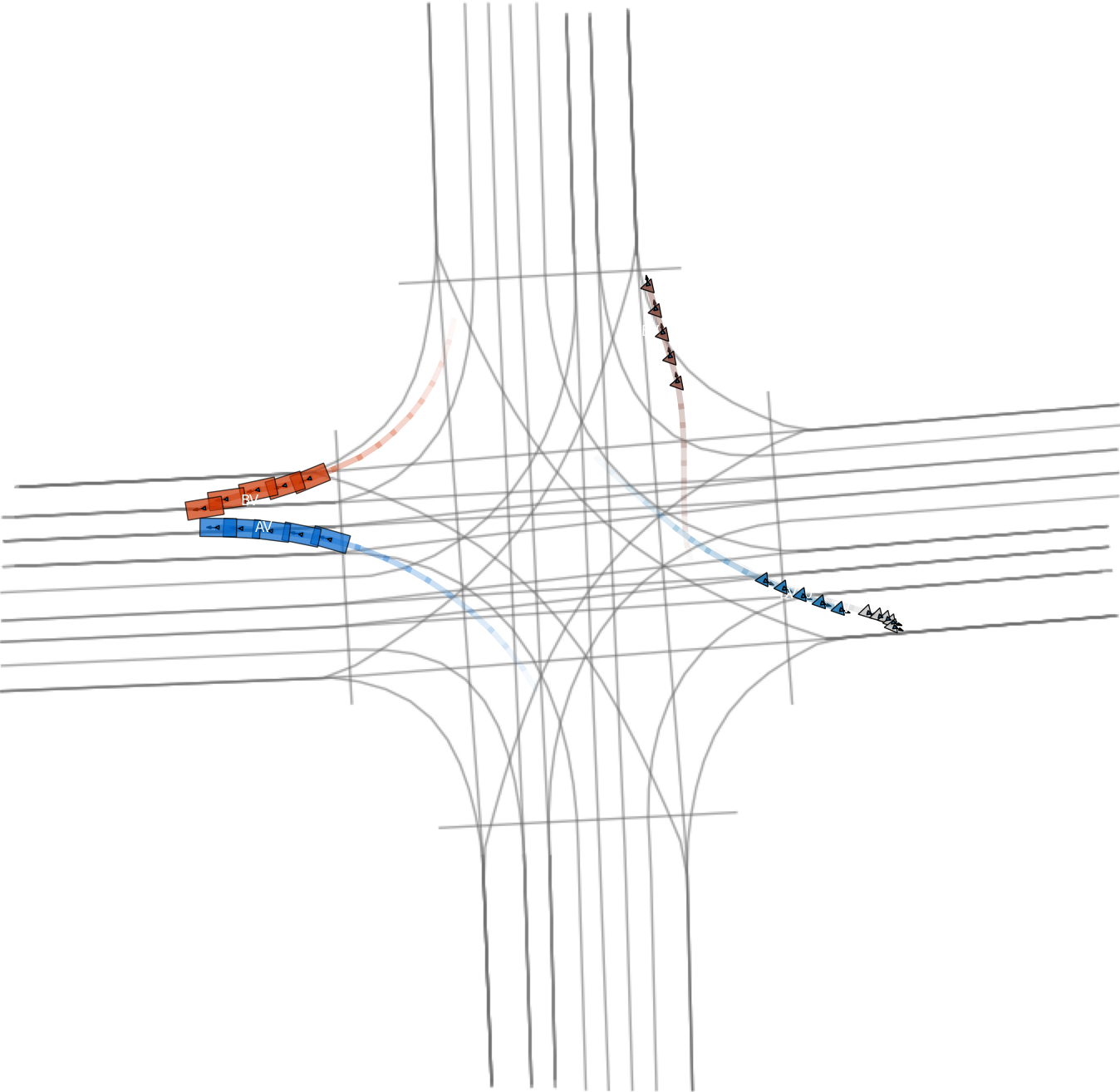}}
\fbox{\includegraphics[width=0.42\textwidth]{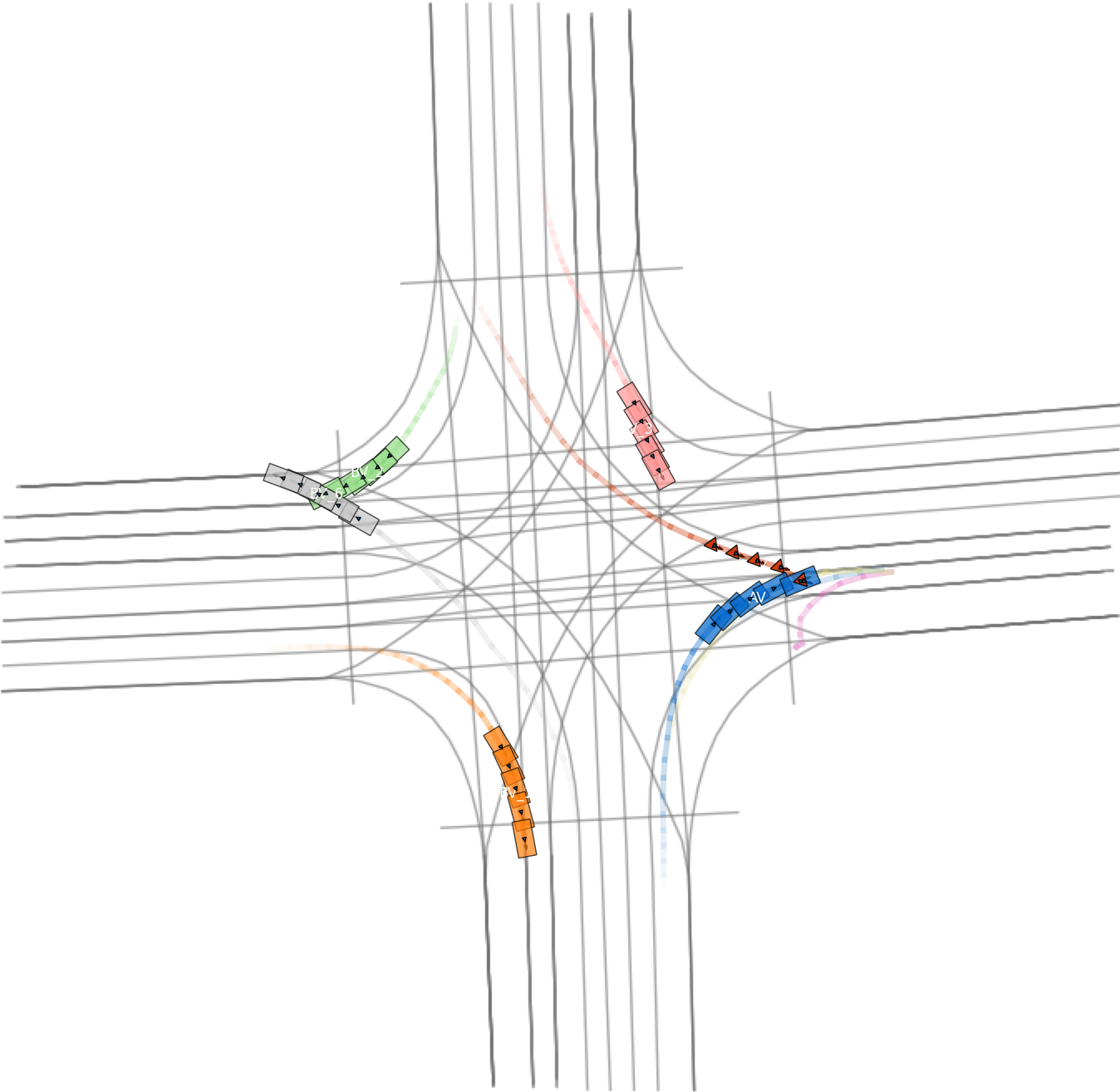}}
\fbox{\includegraphics[width=0.42\textwidth]{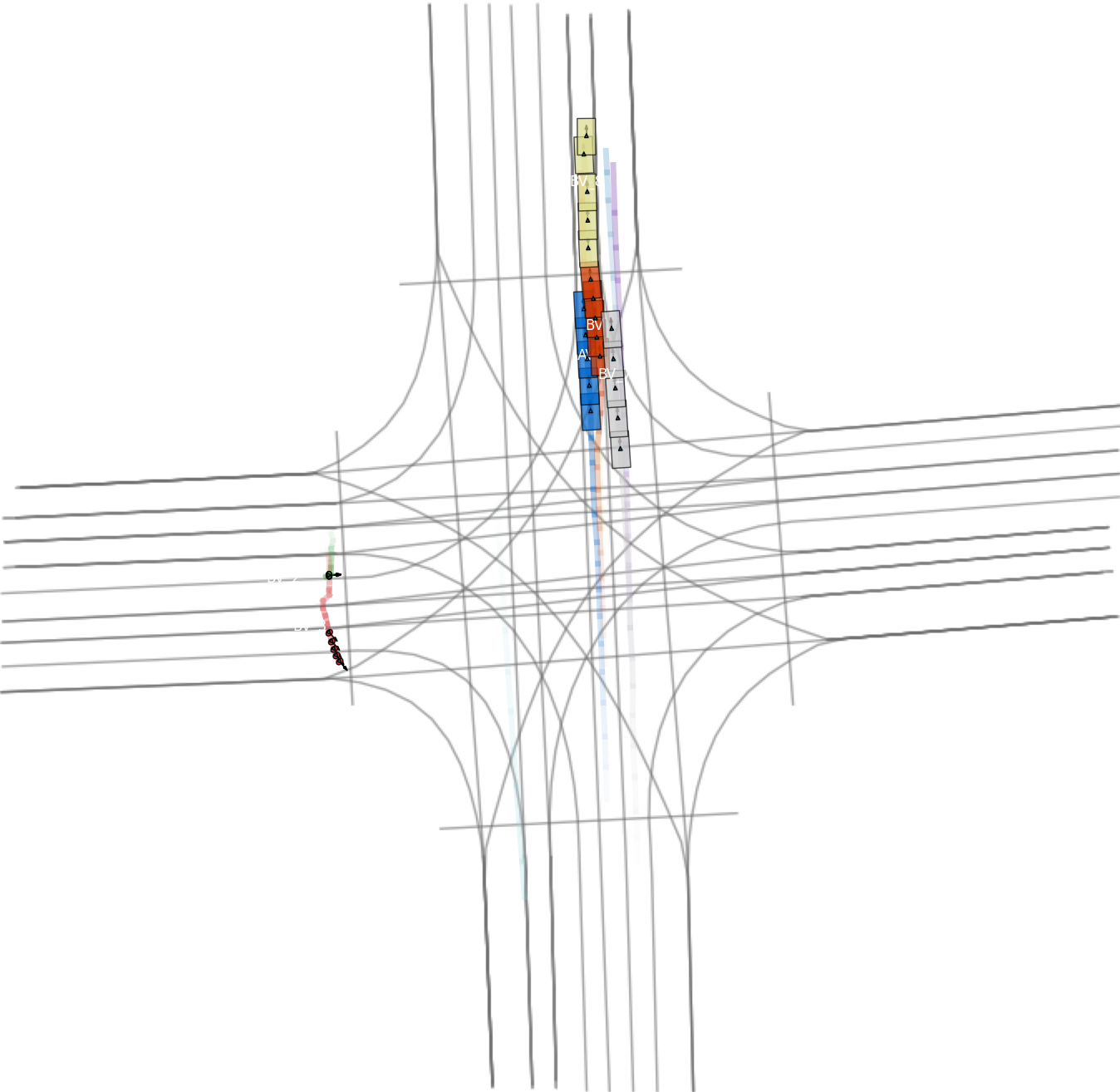}}
  \caption{Representative collision samples involving diverse road users. Vehicles are shown as bounding boxes, bicyclists as triangles, and pedestrians as ellipses. }
  \label{fig:inter_col_samples}
\end{figure*}

Representative qualitative examples are illustrated in Fig.~\ref{fig:inter_col_samples}: (a) A bicyclist executes a left turn while the AV accelerates through a right turn, resulting in a collision on the AV's left side; (b) After passing the intersection, the AV performs a slight left lane change while the BV accelerates straight ahead, leading to a left sideswipe; (c) Several vehicles pass through the intersection: the BV swerves left to avoid a merging lead vehicle, while the AV initiates a slight right lane change from the leftmost lane, culminating in a right sideswipe; (d) The BV turns right into the roadway and subsequently attempts a leftward lane change, while the AV turns left into the same lane, producing a right sideswipe; (e) A bicyclist turns left into the lane as the AV turns right into the same lane, resulting in a frontal collision; (f) As the BV's lead vehicle decelerates, the BV slows down, while the AV simultaneously changes to the right lane, leading to a rear-end collision.

Collectively, these results confirm that CRAG not only increases the likelihood of inducing risky events but also broadens the coverage of intersection scenarios, capturing diverse interaction patterns among vehicles, bicyclists, and pedestrians. This controllability is critical for scaling AV safety tests to more realistic and safety-critical urban conditions.

\subsection{Ablation Study}
\label{subsec:abla}
To assess the effectiveness of the targeted loss, we conduct an ablation in which the probability-based term is removed and the objective is replaced by a mean-squared error (MSE) on latent means. Results are summarized in Table~\ref{tb:ab}. To evaluate the role of the distance metric itself, we also test alternative distributional distances that account for encoder uncertainty, including the Kullback--Leibler (KL) divergence \cite{kullback1951information} and the Wasserstein distance \cite{panaretos2019statistical}, both computed using the estimated variance of the encoded posterior.

As shown in Table~\ref{tb:ab}, removing the probability-based term weakens target-type controllability: TCR drops from 11.0\% (ours) to 9.7\% with MSE and 8.5\% with the Wasserstein loss, even though overall collision frequency (GCR) may increase. The MSE objective compares only latent means and ignores dispersion, providing a coarse but numerically stable signal that exerts weaker alignment pressure toward the target type. Consequently, the resulting feature distribution within the target collision category deviates more substantially from the original distribution. The Wasserstein distance respects distributional geometry (mean and variance) and yields comparable but still lower controllability under the same step budget, along with a lower overall collision rate. Meanwhile, it also produces larger deviations in $\Delta y$ and $\Delta x$ features compared to CRAG. By contrast, the KL objective produces 0 collisions within 30 gradient steps; empirically, its loss decreases slowly and the gradient landscape near the prior is relatively flat, offering weaker guidance than the steeper, more informative Wasserstein gradients under identical optimization settings.

In summary, compared to other distance metrics, combining the probability-based term with MSE loss yields better controllability and a higher success rate in generating targeted collisions while preserving distributional similarity.

\section{Conclusion and Future Work}
\label{sec:conclusion}
This paper introduces a system for generating realistic scenarios that encompass both safe and safety-critical situations. By constructing a structured latent space that bridges the gap between scarce accident data and abundant nominal driving data, the framework enables rapid adaptation to diverse road networks and localized accident records, thereby supporting targeted and scalable evaluation of autonomous driving systems. Experiments demonstrate the effectiveness of the approach in generating controllable safety-critical scenarios and its adaptability to different motion predictor architectures.
Looking ahead, we plan to extend the framework along several directions. First, we will extend latent space modeling to longer-horizon interaction dynamics, leveraging realistic accident datasets and incorporating explicit dynamic constraints to capture complex temporal evolutions. Second, we will explore natural language interfaces through pure text descriptions, making scenario generation more accessible, flexible, and user-friendly. Finally, we aim to integrate diverse AV system models—including both rule-based and end-to-end approaches—to exploit opportunities for collaborative learning and generation.

\bibliographystyle{IEEEtran}
\bibliography{CRAG}

@String(CVPR= {IEEE Conf. Comput. Vis. Pattern Recog.})

@String(ECCV= {Eur. Conf. Comput. Vis.})

@String(CVPR  = {CVPR})

@String(ECCV  = {ECCV})

@article{kingma2013auto,
  title={Auto-encoding variational bayes},
  author={Kingma, Diederik P and Welling, Max},
  journal={arXiv preprint arXiv:1312.6114},
  year={2013}
}

@inproceedings{krajewski2018highd,
  title={The highd dataset: A drone dataset of naturalistic vehicle trajectories on german highways for validation of highly automated driving systems},
  author={Krajewski, Robert and Bock, Julian and Kloeker, Laurent and Eckstein, Lutz},
  booktitle={2018 21st international conference on intelligent transportation systems (ITSC)},
  pages={2118--2125},
  year={2018},
  organization={IEEE}
}

@article{ho2020denoising,
  title={Denoising diffusion probabilistic models},
  author={Ho, Jonathan and Jain, Ajay and Abbeel, Pieter},
  journal={Advances in neural information processing systems},
  volume={33},
  pages={6840--6851},
  year={2020}
}

@article{goodfellow2020generative,
  title={Generative adversarial networks},
  author={Goodfellow, Ian and Pouget-Abadie, Jean and Mirza, Mehdi and Xu, Bing and Warde-Farley, David and Ozair, Sherjil and Courville, Aaron and Bengio, Yoshua},
  journal={Communications of the ACM},
  volume={63},
  number={11},
  pages={139--144},
  year={2020},
  publisher={ACM New York, NY, USA}
}

@inproceedings{caesar2020nuscenes,
  title={nuscenes: A multimodal dataset for autonomous driving},
  author={Caesar, Holger and Bankiti, Varun and Lang, Alex H and Vora, Sourabh and Liong, Venice Erin and Xu, Qiang and Krishnan, Anush and Pan, Yu and Baldan, Giancarlo and Beijbom, Oscar},
  booktitle={Proceedings of the IEEE/CVF conference on computer vision and pattern recognition},
  pages={11621--11631},
  year={2020}
}

@inproceedings{mei2022waymo,
  title={Waymo open dataset: Panoramic video panoptic segmentation},
  author={Mei, Jieru and Zhu, Alex Zihao and Yan, Xinchen and Yan, Hang and Qiao, Siyuan and Chen, Liang-Chieh and Kretzschmar, Henrik},
  booktitle={European Conference on Computer Vision},
  pages={53--72},
  year={2022},
  organization={Springer}
}

@article{liu2024curse,
  title={Curse of rarity for autonomous vehicles},
  author={Liu, Henry X and Feng, Shuo},
  journal={nature communications},
  volume={15},
  number={1},
  pages={4808},
  year={2024},
  publisher={Nature Publishing Group UK London}
}

@article{cho2014learning,
  title={Learning phrase representations using RNN encoder-decoder for statistical machine translation},
  author={Cho, Kyunghyun and Van Merri{\"e}nboer, Bart and Gulcehre, Caglar and Bahdanau, Dzmitry and Bougares, Fethi and Schwenk, Holger and Bengio, Yoshua},
  journal={arXiv preprint arXiv:1406.1078},
  year={2014}
}

@article{feng2023dense,
  title={Dense reinforcement learning for safety validation of autonomous vehicles},
  author={Feng, Shuo and Sun, Haowei and Yan, Xintao and Zhu, Haojie and Zou, Zhengxia and Shen, Shengyin and Liu, Henry X},
  journal={Nature},
  volume={615},
  number={7953},
  pages={620--627},
  year={2023},
  publisher={Nature Publishing Group UK London}
}

@article{fremont2023scenic,
  title={Scenic: A language for scenario specification and data generation},
  author={Fremont, Daniel J and Kim, Edward and Dreossi, Tommaso and Ghosh, Shromona and Yue, Xiangyu and Sangiovanni-Vincentelli, Alberto L and Seshia, Sanjit A},
  journal={Machine Learning},
  volume={112},
  number={10},
  pages={3805--3849},
  year={2023},
  publisher={Springer}
}

@article{yan2023learning,
  title={Learning naturalistic driving environment with statistical realism},
  author={Yan, Xintao and Zou, Zhengxia and Feng, Shuo and Zhu, Haojie and Sun, Haowei and Liu, Henry X},
  journal={Nature communications},
  volume={14},
  number={1},
  pages={2037},
  year={2023},
  publisher={Nature Publishing Group UK London}
}

@inproceedings{higgins2017beta,
  title={beta-vae: Learning basic visual concepts with a constrained variational framework},
  author={Higgins, Irina and Matthey, Loic and Pal, Arka and Burgess, Christopher and Glorot, Xavier and Botvinick, Matthew and Mohamed, Shakir and Lerchner, Alexander},
  booktitle={International conference on learning representations},
  year={2017}
}

@article{chen2024data,
  title={Data-driven Traffic Simulation: A Comprehensive Review},
  author={Chen, Di and Zhu, Meixin and Yang, Hao and Wang, Xuesong and Wang, Yinhai},
  journal={IEEE Transactions on Intelligent Vehicles},
  year={2024},
  publisher={IEEE}
}

@inproceedings{arief2018accelerated,
  title={An accelerated approach to safely and efficiently test pre-production autonomous vehicles on public streets},
  author={Arief, Mansur and Glynn, Peter and Zhao, Ding},
  booktitle={2018 21st International Conference on Intelligent Transportation Systems (ITSC)},
  pages={2006--2011},
  year={2018},
  organization={IEEE}
}

@article{li2024scenarionet,
  title={Scenarionet: Open-source platform for large-scale traffic scenario simulation and modeling},
  author={Li, Quanyi and Peng, Zhenghao Mark and Feng, Lan and Liu, Zhizheng and Duan, Chenda and Mo, Wenjie and Zhou, Bolei},
  journal={Advances in neural information processing systems},
  volume={36},
  year={2024}
}

@article{he2024knowmoformer,
  title     = {KnowMoformer: Knowledge-Conditioned Motion Transformer for Controllable Traffic Scenario Simulation},
  author    = {He, Honglin and Li, Shu and Yang, Jingxuan and He, Linxuan and Zhang, Yi and Lu, Qiujing and Feng, Shuo},
  journal = {Proceedings of the IEEE/CVF Conference on Computer Vision and Pattern Recognition (CVPR)},
  year      = {2024}, 
  note      = {CVPR 2024},
}

@article{panaretos2019statistical,
  title={Statistical aspects of Wasserstein distances},
  author={Panaretos, Victor M and Zemel, Yoav},
  journal={Annual review of statistics and its application},
  volume={6},
  number={1},
  pages={405--431},
  year={2019},
  publisher={Annual Reviews}
}

@article{lin2025intersectionde,
  title={IntersectioNDE: Learning Complex Urban Traffic Dynamics based on Interaction Decoupling Strategy},
  author={Lin, Enli and Yang, Ziyuan and Lu, Qiujing and Hu, Jianming and Feng, Shuo},
  journal={arXiv preprint arXiv:2510.11534},
  year={2025}
}

@article{kullback1951information,
  title={On information and sufficiency},
  author={Kullback, Solomon and Leibler, Richard A},
  journal={The annals of mathematical statistics},
  volume={22},
  number={1},
  pages={79--86},
  year={1951},
  publisher={JSTOR}
}

@article{gulino2024waymax,
  title={Waymax: An accelerated, data-driven simulator for large-scale autonomous driving research},
  author={Gulino, Cole and Fu, Justin and Luo, Wenjie and Tucker, George and Bronstein, Eli and Lu, Yiren and Harb, Jean and Pan, Xinlei and Wang, Yan and Chen, Xiangyu and others},
  journal={Advances in Neural Information Processing Systems},
  volume={36},
  year={2024}
}

@article{liu2024towards,
  title={Towards Interactive Autonomous Vehicle Testing: Vehicle-Under-Test-Centered Traffic Simulation},
  author={Liu, Yiru and Zhao, Xiaocong and Sun, Jian},
  journal={arXiv preprint arXiv:2406.02860},
  year={2024}
}

@article{ren2025intelligent,
  title={Intelligent testing environment generation for autonomous vehicles with implicit distributions of traffic behaviors},
  author={Ren, Kun and Yang, Jingxuan and Lu, Qiujing and Zhang, Yi and Hu, Jianming and Feng, Shuo},
  journal={Transportation Research Part C: Emerging Technologies},
  volume={174},
  pages={105106},
  year={2025},
  publisher={Elsevier}
}

@article{caesar2021nuplan,
  title={nuplan: A closed-loop ml-based planning benchmark for autonomous vehicles},
  author={Caesar, Holger and Kabzan, Juraj and Tan, Kok Seang and Fong, Whye Kit and Wolff, Eric and Lang, Alex and Fletcher, Luke and Beijbom, Oscar and Omari, Sammy},
  journal={arXiv preprint arXiv:2106.11810},
  year={2021}
}

@article{montali2023waymo,
  title={The waymo open sim agents challenge},
  author={Montali, Nico and Lambert, John and Mougin, Paul and Kuefler, Alex and Rhinehart, Nicholas and Li, Michelle and Gulino, Cole and Emrich, Tristan and Yang, Zoey and Whiteson, Shimon and others},
  journal={Advances in Neural Information Processing Systems},
  volume={36},
  pages={59151--59171},
  year={2023}
}

@article{vinitsky2022nocturne,
  title={Nocturne: a scalable driving benchmark for bringing multi-agent learning one step closer to the real world},
  author={Vinitsky, Eugene and Lichtl{\'e}, Nathan and Yang, Xiaomeng and Amos, Brandon and Foerster, Jakob},
  journal={Advances in Neural Information Processing Systems},
  volume={35},
  pages={3962--3974},
  year={2022}
}

@article{li2022metadrive,
  title={Metadrive: Composing diverse driving scenarios for generalizable reinforcement learning},
  author={Li, Quanyi and Peng, Zhenghao and Feng, Lan and Zhang, Qihang and Xue, Zhenghai and Zhou, Bolei},
  journal={IEEE transactions on pattern analysis and machine intelligence},
  volume={45},
  number={3},
  pages={3461--3475},
  year={2022},
  publisher={IEEE}
}

@inproceedings{yan2024street,
  title={Street gaussians: Modeling dynamic urban scenes with gaussian splatting},
  author={Yan, Yunzhi and Lin, Haotong and Zhou, Chenxu and Wang, Weijie and Sun, Haiyang and Zhan, Kun and Lang, Xianpeng and Zhou, Xiaowei and Peng, Sida},
  booktitle={European Conference on Computer Vision},
  pages={156--173},
  year={2024},
  organization={Springer}
}

@article{guo2023streetsurf,
  title={Streetsurf: Extending multi-view implicit surface reconstruction to street views},
  author={Guo, Jianfei and Deng, Nianchen and Li, Xinyang and Bai, Yeqi and Shi, Botian and Wang, Chiyu and Ding, Chenjing and Wang, Dongliang and Li, Yikang},
  journal={arXiv preprint arXiv:2306.04988},
  year={2023}
}

@misc{luracl,
  title={RACL: Risk Aware Closed-Loop Agent Simulation with High Fidelity},
  author={Lu, Qiujing and Bai, Ruoxuan and Li, Shu and He, Honglin and Feng, Shuo},
  url={https://agents4ad.github.io/assets/cvpr2024/papers/30.pdf}
}

@inproceedings{zhang2023cat,
  title={Cat: Closed-loop adversarial training for safe end-to-end driving},
  author={Zhang, Linrui and Peng, Zhenghao and Li, Quanyi and Zhou, Bolei},
  booktitle={Conference on Robot Learning},
  pages={2357--2372},
  year={2023},
  organization={PMLR}
}

@article{lu2024multimodal,
  title={Multimodal large language model driven scenario testing for autonomous vehicles},
  author={Lu, Qiujing and Wang, Xuanhan and Jiang, Yiwei and Zhao, Guangming and Ma, Mingyue and Feng, Shuo},
  journal={arXiv preprint arXiv:2409.06450},
  year={2024}
}

@article{lu2024realistic,
  title={Realistic corner case generation for autonomous vehicles with multimodal large language model},
  author={Lu, Qiujing and Ma, Meng and Dai, Ximiao and Wang, Xuanhan and Feng, Shuo},
  journal={arXiv preprint arXiv:2412.00243},
  year={2024}
}

@article{aasi2024generating,
  title={Generating out-of-distribution scenarios using language models},
  author={Aasi, Erfan and Nguyen, Phat and Sreeram, Shiva and Rosman, Guy and Karaman, Sertac and Rus, Daniela},
  journal={arXiv preprint arXiv:2411.16554},
  year={2024}
}

@article{laugier2011probabilistic,
  title={Probabilistic analysis of dynamic scenes and collision risks assessment to improve driving safety},
  author={Laugier, Christian and Paromtchik, Igor E and Perrollaz, Mathias and Yong, Mao and Yoder, John-David and Tay, Christopher and Mekhnacha, Kamel and N{\`e}gre, Amaury},
  journal={IEEE Intelligent Transportation Systems Magazine},
  volume={3},
  number={4},
  pages={4--19},
  year={2011},
  publisher={IEEE}
}

@article{chen2020driving,
  title={Driving maneuvers prediction based autonomous driving control by deep Monte Carlo tree search},
  author={Chen, Jienan and Zhang, Cong and Luo, Jinting and Xie, Junfei and Wan, Yan},
  journal={IEEE transactions on vehicular technology},
  volume={69},
  number={7},
  pages={7146--7158},
  year={2020},
  publisher={IEEE}
}

@article{aoude2012driver,
  title={Driver behavior classification at intersections and validation on large naturalistic data set},
  author={Aoude, Georges S and Desaraju, Vishnu R and Stephens, Lauren H and How, Jonathan P},
  journal={IEEE Transactions on Intelligent Transportation Systems},
  volume={13},
  number={2},
  pages={724--736},
  year={2012},
  publisher={IEEE}
}

@inproceedings{Rana_Malhi_2021, title={Building Safer Autonomous Agents by Leveraging Risky Driving Behavior Knowledge}, booktitle={2021 International Conference on Communications, Computing, Cybersecurity, and Informatics (CCCI)}, author={Rana, Ashish and Malhi, Avleen}, year={2021}, month=oct, pages={1--6} }

@inproceedings{Althoff_Lutz_2018, title={Automatic Generation of Safety-Critical Test Scenarios for Collision Avoidance of Road Vehicles}, ISSN={1931-0587}, DOI={10.1109/IVS.2018.8500374}, booktitle={2018 IEEE Intelligent Vehicles Symposium (IV)}, author={Althoff, Matthias and Lutz, Sebastian}, year={2018}, month=jun, pages={1326--1333} }

@article{Scanlon_Kusano_Daniel_Alderson_Ogle_Victor_2021, title={Waymo simulated driving behavior in reconstructed fatal crashes within an autonomous vehicle operating domain}, volume={163}, ISSN={0001-4575}, DOI={10.1016/j.aap.2021.106454}, journal={Accident Analysis \& Prevention}, author={Scanlon, John M. and Kusano, Kristofer D. and Daniel, Tom and Alderson, Christopher and Ogle, Alexander and Victor, Trent}, year={2021}, month=dec, pages={106454} }

@article{Ding_Xu_Arief_Lin_Li_Zhao_2023, title={A Survey on Safety-Critical Driving Scenario Generation--A Methodological Perspective}, volume={24}, ISSN={1558-0016}, DOI={10.1109/TITS.2023.3259322}, number={7}, journal={IEEE Transactions on Intelligent Transportation Systems}, author={Ding, Wenhao and Xu, Chejian and Arief, Mansur and Lin, Haohong and Li, Bo and Zhao, Ding}, year={2023}, month=jul, pages={6971--6988} }

@inproceedings{Hanselmann_Renz_Chitta_Bhattacharyya_Geiger_2022, address={Cham}, title={KING: Generating Safety-Critical Driving Scenarios for Robust Imitation via Kinematics Gradients}, ISBN={978-3-031-19839-7}, DOI={10.1007/978-3-031-19839-7_20}, booktitle={Computer Vision -- ECCV 2022}, publisher={Springer Nature Switzerland}, author={Hanselmann, Niklas and Renz, Katrin and Chitta, Kashyap and Bhattacharyya, Apratim and Geiger, Andreas}, editor={Avidan, Shai and Brostow, Gabriel and CissÃ©, Moustapha and Farinella, Giovanni Maria and Hassner, Tal}, year={2022}, pages={335--352}, language={en} }

@inproceedings{Wang_Pun_Tu_Manivasagam_Sadat_Casas_Ren_Urtasun_2021, title={AdvSim: Generating Safety-Critical Scenarios for Self-Driving Vehicles}, author={Wang, Jingkang and Pun, Ava and Tu, James and Manivasagam, Sivabalan and Sadat, Abbas and Casas, Sergio and Ren, Mengye and Urtasun, Raquel}, year={2021}, pages={9909--9918}, language={en} }

@inproceedings{Rempe_Philion_Guibas_Fidler_Litany_2022, address={New Orleans, LA, USA}, title={Generating Useful Accident-Prone Driving Scenarios via a Learned Traffic Prior}, rights={https://doi.org/10.15223/policy-029}, ISBN={978-1-66546-946-3}, DOI={10.1109/CVPR52688.2022.01679}, booktitle={2022 IEEE/CVF Conference on Computer Vision and Pattern Recognition (CVPR)}, publisher={IEEE}, author={Rempe, Davis and Philion, Jonah and Guibas, Leonidas J. and Fidler, Sanja and Litany, Or}, year={2022}, month=jun, pages={17284--17294}, language={en} }

@inproceedings{Zhong_Rempe_Xu_Chen_Veer_Che_Ray_Pavone_2023, title={Guided Conditional Diffusion for Controllable Traffic Simulation}, DOI={10.1109/ICRA48891.2023.10161463}, booktitle={2023 IEEE International Conference on Robotics and Automation (ICRA)}, author={Zhong, Ziyuan and Rempe, Davis and Xu, Danfei and Chen, Yuxiao and Veer, Sushant and Che, Tong and Ray, Baishakhi and Pavone, Marco}, year={2023}, month=may, pages={3560--3566} }

@article{chang2023controllable,
  title={Controllable Safety-Critical Closed-loop Traffic Simulation via Guided Diffusion},
  author={Chang, Wei-Jer and Pittaluga, Francesco and Tomizuka, Masayoshi and Zhan, Wei and Chandraker, Manmohan},
  journal={arXiv preprint arXiv:2401.00391},
  year={2023}
}

@inproceedings{Feng_Li_Peng_Tan_Zhou_2023, title={TrafficGen: Learning to Generate Diverse and Realistic Traffic Scenarios},booktitle={2023 IEEE International Conference on Robotics and Automation (ICRA)}, author={Feng, Lan and Li, Quanyi and Peng, Zhenghao and Tan, Shuhan and Zhou, Bolei}, year={2023}, month=may, pages={3567--3575} }

@inproceedings{Igl_Kim_Kuefler_Mougin_Shah_Shiarlis_Anguelov_Palatucci_White_Whiteson_2022, title={Symphony: Learning Realistic and Diverse Agents for Autonomous Driving Simulation},DOI={10.1109/ICRA46639.2022.9811990}, booktitle={2022 International Conference on Robotics and Automation (ICRA)}, author={Igl, Maximilian and Kim, Daewoo and Kuefler, Alex and Mougin, Paul and Shah, Punit and Shiarlis, Kyriacos and Anguelov, Dragomir and Palatucci, Mark and White, Brandyn and Whiteson, Shimon}, year={2022}, month=may, pages={2445--2451} }

@article{Wang_Zhao_Yi_2023, title={Multiverse Transformer: 1st Place Solution for Waymo Open Sim Agents Challenge 2023}, abstractNote={This technical report presents our 1st place solution for the Waymo Open Sim Agents Challenge (WOSAC) 2023. Our proposed MultiVerse Transformer for Agent simulation (MVTA) effectively leverages transformer-based motion prediction approaches, and is tailored for closed-loop simulation of agents. In order to produce simulations with a high degree of realism, we design novel training and sampling methods, and implement a receding horizon prediction mechanism. In addition, we introduce a variable-length history aggregation method to mitigate the compounding error that can arise during closed-loop autoregressive execution. On the WOSAC, our MVTA and its enhanced version MVTE reach a realism meta-metric of 0.5091 and 0.5168, respectively, outperforming all the other methods on the leaderboard.}, year={2023},  number={arXiv:2306.11868}, publisher={arXiv}, author={Wang, Yu and Zhao, Tiebiao and Yi, Fan}}

@inproceedings{Suo_Regalado_Casas_Urtasun_2021, title={TrafficSim: Learning To Simulate Realistic Multi-Agent Behaviors}, author={Suo, Simon and Regalado, Sebastian and Casas, Sergio and Urtasun, Raquel}, year={2021}, pages={10400--10409}, language={en} }

@article{Treiber_Hennecke_Helbing_2000, title={Congested traffic states in empirical observations and microscopic simulations}, volume={62}, DOI={10.1103/PhysRevE.62.1805}, number={2}, journal={Physical Review E}, publisher={American Physical Society}, author={Treiber, Martin and Hennecke, Ansgar and Helbing, Dirk}, year={2000}, month=aug, pages={1805--1824} }

@article{Kesting_Treiber_Helbing_2007, title={General lane-changing model MOBIL for car-following models}, volume={1999}, ISBN={0361-1981}, number={1}, journal={Transportation Research Record}, publisher={SAGE Publications Sage CA: Los Angeles, CA}, author={Kesting, Arne and Treiber, Martin and Helbing, Dirk}, year={2007}, pages={86--94} }

\newpage
\section{Biography Section}
\begin{IEEEbiography}[{\includegraphics[width=1in,height=1.25in,clip,keepaspectratio]{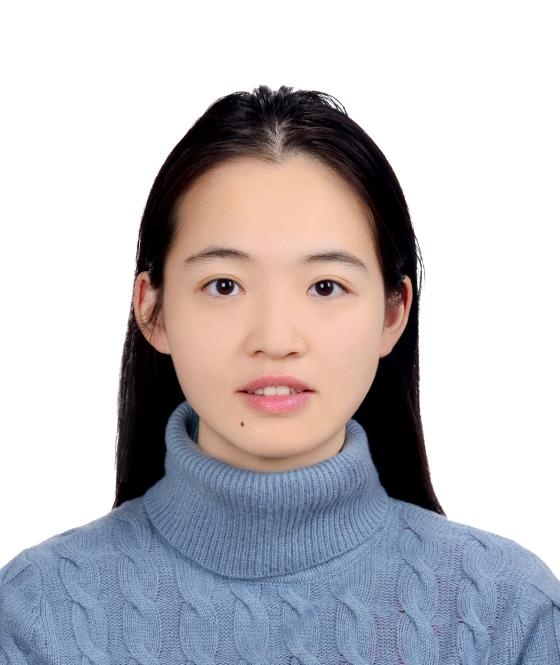}}]{Qiujing Lu} received the B.S. degree in automation from Tsinghua University, Beijing, China, in 2015, and the Ph.D. degree in electrical and computer engineering from the University of California at Los Angeles, Los Angeles, CA, USA, in 2022. She is currently a postdoctoral research fellow in the Department of Automation at Tsinghua University. Her research focuses on generative models and safe autonomous driving.
\end{IEEEbiography}
\begin{IEEEbiography}[{\includegraphics[width=1in,height=1.25in,clip,keepaspectratio]{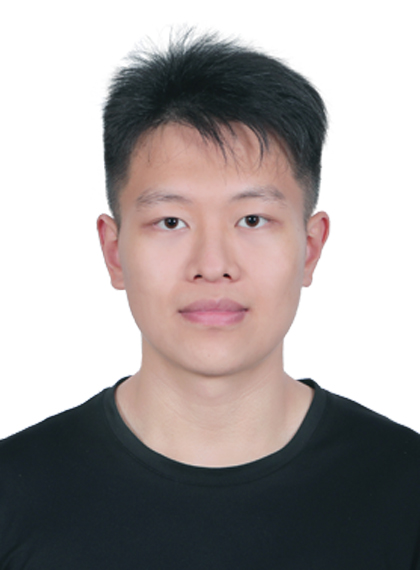}}]{Xuanhan Wang} received the bachelor's degree from Tsinghua University, Beijing, in 2025. He is currently pursuing the Ph.D. degree in the Department of Automation, Tsinghua University, under the supervision of Prof. Shuo Feng. His research interests include simulation-based testing, safety-critical scenario generation, and safety evaluation of AI systems.
\end{IEEEbiography}

\begin{IEEEbiography}[{\includegraphics[width=1in,height=1.25in,clip,keepaspectratio]{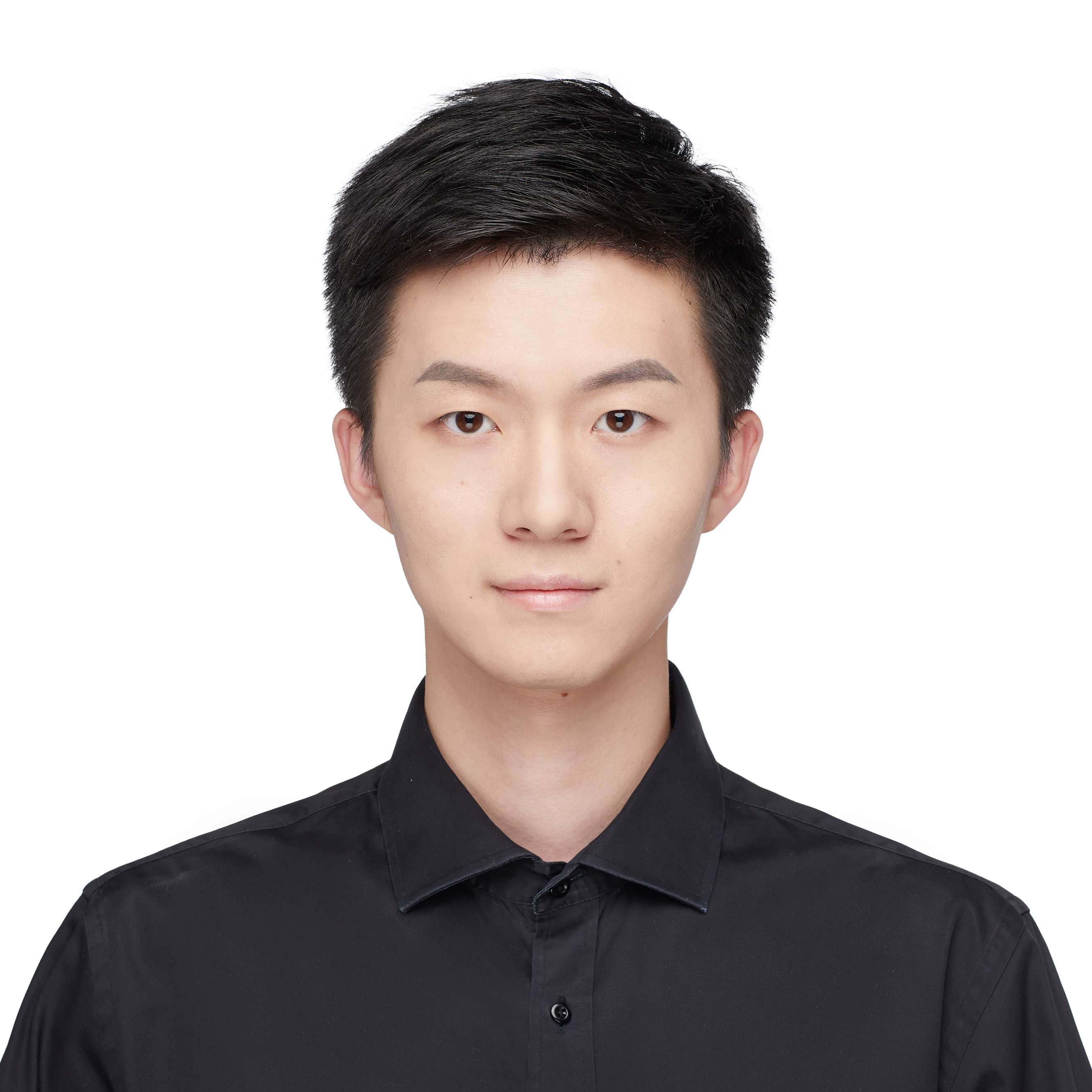}}]{Runze Yuan}
received the bachelor’s degree from Tsinghua University, Beijing, in 2017, and the master’s degree and Ph.D. degree from the University of Hawaii at Manoa, Honolulu, in 2020 and 2023, respectively. He is currently working as a Postdoc of the Department of Automation, Tsinghua University. His main research interests include microscopic traffic flow modeling, artificial intelligence in transportation,
traffic flow detection and control, and intelligent transportation systems.
\end{IEEEbiography}
\begin{IEEEbiography}[{\includegraphics[width=1in,height=1.25in,clip,keepaspectratio]{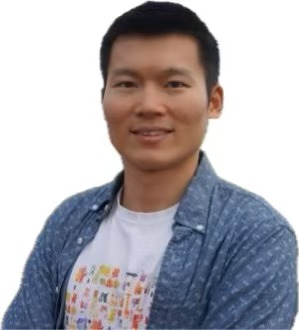}}]{Wei Lu} received his Ph.D degree in Transportation Engineering and Computational Science from The University of Tennessee. He has worked at the Oak Ridge National Lab and multiple private companies in the autonomous vehicle industry. His interests concentrate on AI and robotics safety, simulation, and evaluation.
\end{IEEEbiography}
\begin{IEEEbiography}[{\includegraphics[width=1in,height=1.25in,clip,keepaspectratio]{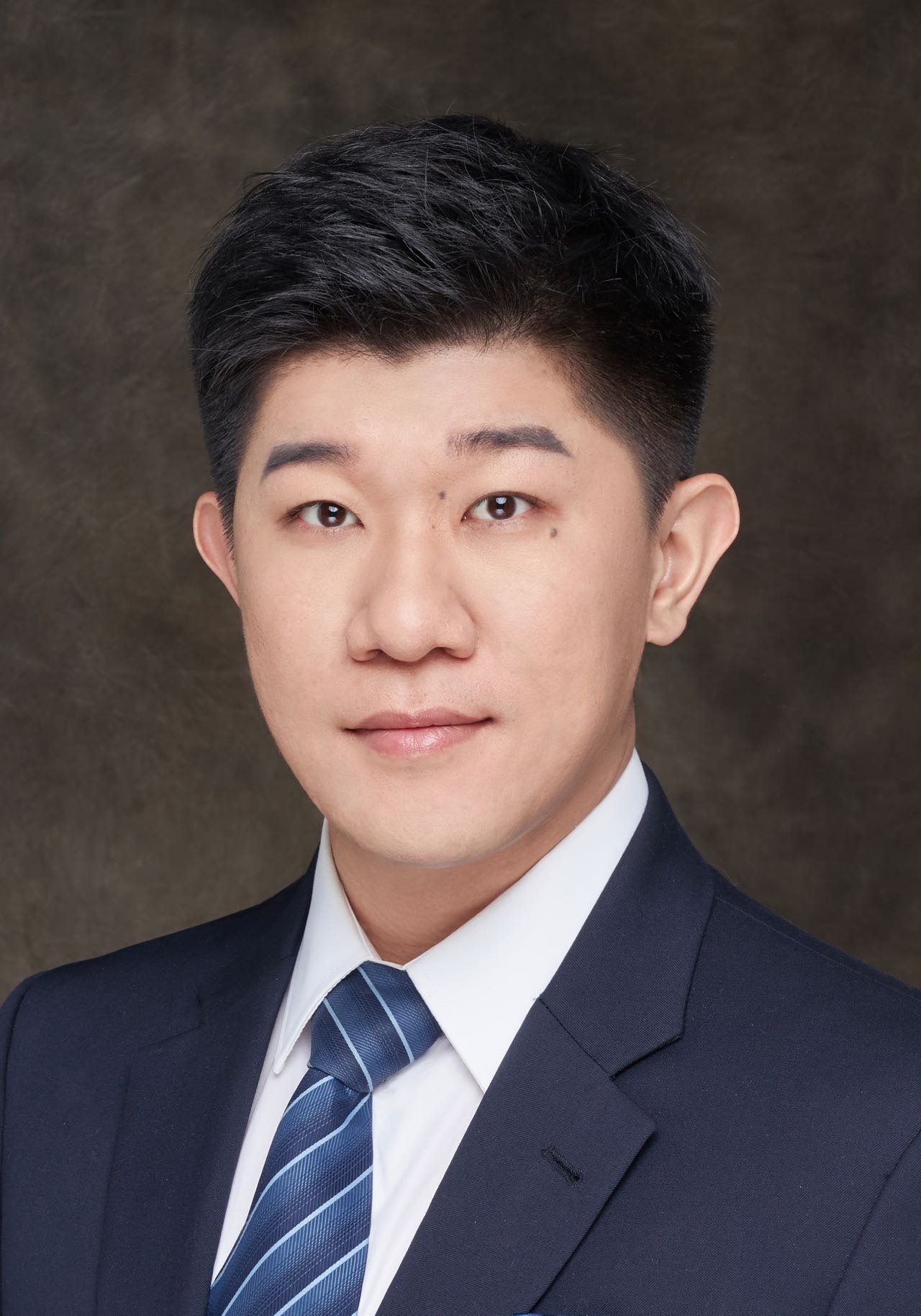}}]{Xinyi Gong} received the Ph.D degree in control theory and control engineering from the Institute of Automation (IA), Chinese Academy of Sciences (CAS), Beijing, China, in 2019. He is currently a Professor with the Space Information Research Institute, Hangzhou Dianzi University (HDU), Hangzhou, China. His research interests include computer vision, image processing, pattern recognition, and machine learning.
\end{IEEEbiography}
\begin{IEEEbiography}[{\includegraphics[width=1in,height=1.25in,clip,keepaspectratio]{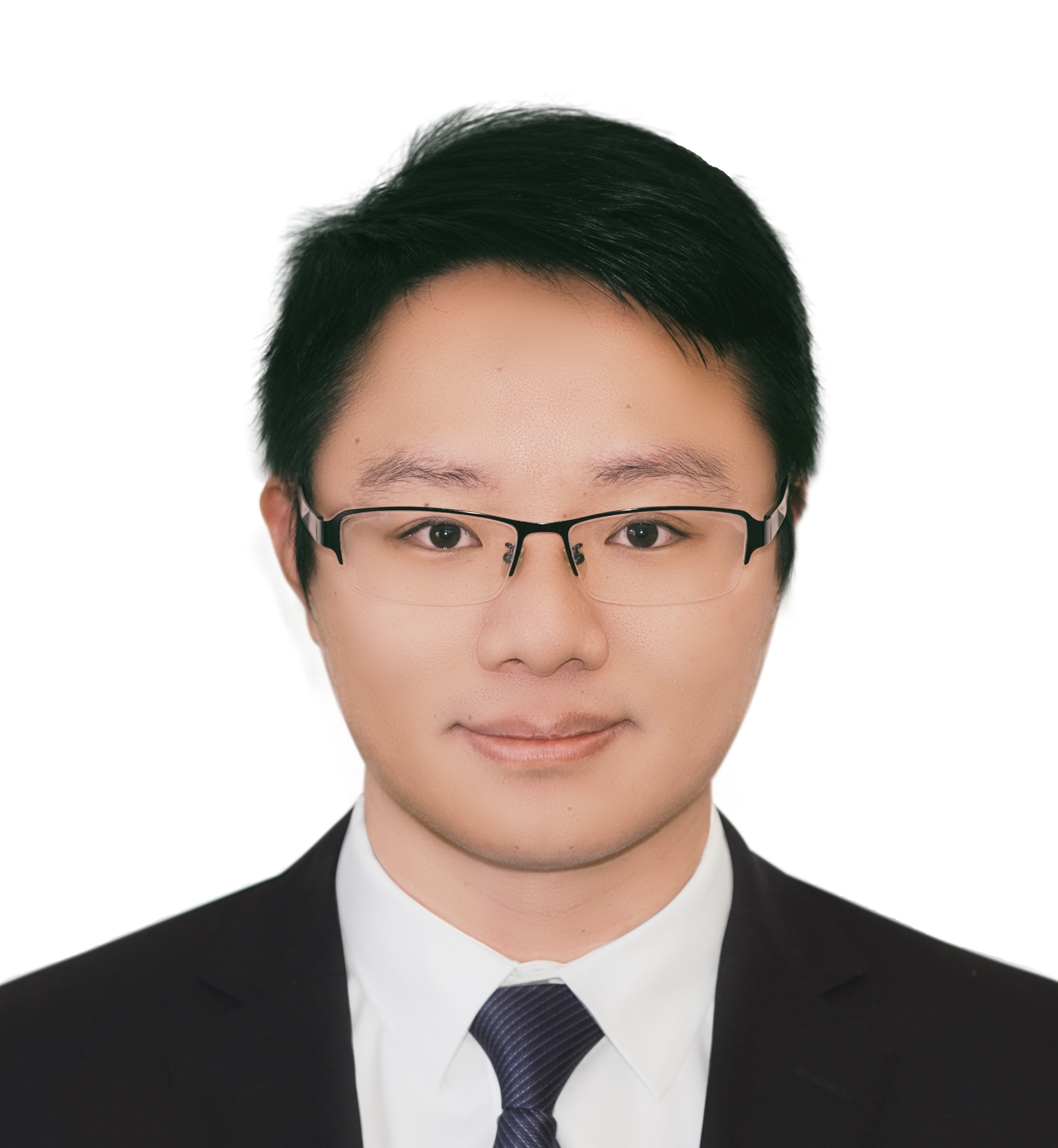}}]{Shuo Feng} (Member, IEEE) received the bachelor’s and Ph.D. degrees in the Department of Automation at Tsinghua University, China, in 2014 and 2019,
respectively. He was a postdoctoral research fellow in the Department of Civil and Environmental Engineering and also an Assistant Research Scientist at the University of Michigan Transportation Research Institute (UMTRI) at the University of Michigan, Ann Arbor. He is currently an Assistant Professor in the Department of Automation at Tsinghua University. His research interests lie in the development and validation of safety-critical machine learning, particularly for connected and automated vehicles. He was a recipient of the Best Ph.D. Dissertation Award from the IEEE Intelligent Transportation Systems Society in 2020 and the ITS Best Paper Award from the INFORMS TSL society in 2021. He is an Associate Editor of the IEEE TRANSACTIONS ON INTELLIGENT VEHICLES and an Academic Editor of the Automotive Innovation.
\end{IEEEbiography}

\vfill

\end{document}